%% file: ECHO-arxiv.tex
\documentclass[sigconf,screen,nonacm]{acmart}

\AtBeginDocument{%
  }

\setcopyright{none}
\usepackage{array}
\usepackage{multirow}
\usepackage{siunitx}
\usepackage{etoolbox}
\usepackage{makecell}
\usepackage{tabularx}
\usepackage{placeins}
\usepackage[ruled,vlined]{algorithm2e} % ACM 官方推荐使用的算法包

\usepackage{listings}
\usepackage[most]{tcolorbox} 

\lstdefinestyle{jsonstyle}{
  basicstyle=\ttfamily\footnotesize,
  columns=fullflexible,
  keepspaces=true,
  breaklines=true,
  breakatwhitespace=false,
  showstringspaces=false,
  frame=single,       % 注意：这个边框会在下面被 inputjsonbox 强制取消
  framerule=0.4pt,
  xleftmargin=0.2em,
  xrightmargin=0.2em
}

\definecolor{academicBlue}{RGB}{44, 62, 80}
\definecolor{lightBack}{RGB}{245, 247, 250}
\definecolor{papergray}{gray}{0.95} % 预定义安全背景色

\newtcolorbox{codeboxframe}[1]{
    enhanced,breakable,
    title={#1},
    colback=papergray,colframe=black!70,
    colbacktitle=papergray,coltitle=black,
    fonttitle=\bfseries\small,
    attach boxed title to top left={xshift=4mm, yshift*=-\tcboxedtitleheight/2},
    boxed title style={colback=papergray,colframe=black!70},
    sharp corners,rounded corners=southeast,rounded corners=southwest,
    top=1mm,bottom=1mm,left=1mm,right=1mm,arc=0mm,boxsep=1mm
}

\newcommand{\inputjsonbox}[3][]{%
    \begin{codeboxframe}{#3}
        \lstinputlisting[
            style=jsonstyle,frame=none,backgroundcolor={},
            language=Python,
            basicstyle=\ttfamily\small,
            stringstyle={},keywordstyle=\bfseries,commentstyle=\itshape,
            morekeywords={true,false,null,event_type,trigger,role,text,bounding_box},
            breaklines=true,showstringspaces=false,columns=fullflexible,
            firstline=1,#1
        ]{#2}
    \end{codeboxframe}
}

\newcommand{\inputtextbox}[3][]{%
    \begin{codeboxframe}{#3}
        \lstinputlisting[
            style=jsonstyle,frame=none,backgroundcolor={},
            language={},
            basicstyle=\ttfamily\small,
            breaklines=true,columns=fullflexible,#1
        ]{#2}
    \end{codeboxframe}
}

\newcommand{\best}[1]{\textbf{#1}}
\newcommand{\second}[1]{\underline{#1}}
\newcommand{\NA}{--}
\newrobustcmd{\B}{\fontseries{b}\selectfont}

\begin{document}

\title{ECHO: Event-Centric Hypergraph Operations via Multi-Agent Collaboration for Multimedia Event Extraction}

%% 1. Hailong Chu
\author{Hailong Chu}
\authornote{These authors contributed equally to this work.}
\email{hailongchu@bupt.edu.cn}
\affiliation{%
  \institution{Beijing University of Posts and Telecommunications}
  \city{Beijing}
  \country{China}
}

%% 2. Hongbing Li
\author{Hongbing Li}
\authornotemark[1]
\email{hbl@bupt.edu.cn}
\affiliation{%
  \institution{Beijing University of Posts and Telecommunications}
  \city{Beijing}
  \country{China}
}

%% 3. Yunlong Chu
\author{Yunlong Chu}
\authornotemark[1]
\email{cyl2024245030@tju.edu.cn}
\affiliation{%
  \institution{Tianjin University}
  \city{Tianjin}
  \country{China}
}

%% 4. Shutai Huang
\author{Shutai Huang}
\email{shutaihuang@bupt.edu.cn}
\affiliation{%
  \institution{Beijing University of Posts and Telecommunications}
  \city{Beijing}
  \country{China}
}

%% 5. Xingyue Zhang
\author{Xingyue Zhang}
\email{xingyuezhang@bupt.edu.cn}
\affiliation{%
  \institution{Beijing University of Posts and Telecommunications}
  \city{Beijing}
  \country{China}
}

%% 6. Tinghe Yan
\author{Tinghe Yan}
\email{tingheyan@stu.cqupt.edu.cn}
\affiliation{%
  \institution{Chongqing University of Posts and Telecommunications}
  \city{Chongqing}
  \country{China}
}

%% 7. Jinsong Zhang
\author{Jinsong Zhang}
\email{2024211124@stu.hit.edu.cn}
\affiliation{%
  \institution{Harbin Institute of Technology}
  \city{Harbin}
  \country{China}
}

%% 8. Shuo Zhang
\author{Shuo Zhang}
\authornote{Corresponding authors.}
\email{shuoz@bupt.edu.cn}
\affiliation{%
  \institution{Beijing University of Posts and Telecommunications}
  \city{Beijing}
  \country{China}
}

%% 9. Lei Li
\author{Lei Li}
\authornotemark[2]
\email{leili@bupt.edu.cn}
\affiliation{%
  \institution{Beijing University of Posts and Telecommunications}
  \city{Beijing}
  \country{China}
}

\renewcommand{\shortauthors}{Hailong Chu et al.}


\begin{abstract}
  % Multimedia Event Extraction (M2E2) involves extracting structured event records from both textual and visual content. Existing approaches, ranging from specialized architectures to direct Large Language Model (LLM) prompting, typically rely on a linear, end-to-end generation and thus suffer from cascading errors: early cross-modal misalignments often corrupt downstream role assignment under strict grounding constraints. We propose \textbf{ECHO} (\textbf{E}vent-\textbf{C}entric \textbf{H}ypergraph \textbf{O}perations), a multi-agent framework that iteratively refines a shared \textbf{Multimedia Event Hypergraph (MEHG)}, which serves as an explicit intermediate structure for multimodal event hypotheses. Unlike dialogue-centric frameworks, ECHO coordinates specialized agents by applying atomic hypergraph operations to the MEHG. Furthermore, we introduce a Link-then-Bind strategy that enforces deferred commitment: agents first identify relevant arguments and only then determine their precise roles, mitigating incorrect grounding and limiting error propagation. Extensive experiments on the M2E2 benchmark show that ECHO significantly outperforms the state-of-the-art (SOTA): with Qwen3-32B, it achieves a 7.3\% and 15.5\% improvement in average event mention and argument role F1, respectively.
  \input{body/0-abs}
\end{abstract}





\maketitle

\begin{figure}[t]
    \centering
    \includegraphics[width=1.0\linewidth]{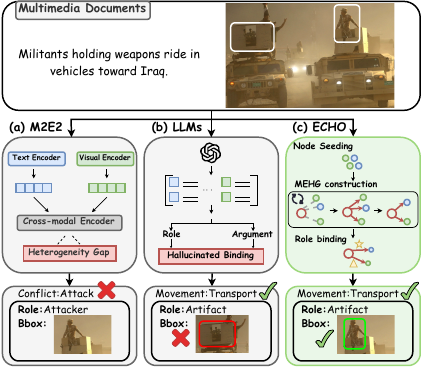}
    \caption{Illustrative comparison of M2E2 paradigms on the same multimedia document. (a) Traditional encoder-based systems rely on implicit event-level structure and may suffer from the heterogeneity gap. (b) Directly prompted LLMs/LVLMs entangle event identification, argument linking, and role assignment in one-shot generation, which can lead to hallucinated bindings. (c) ECHO performs iterative multi-agent revision over an explicit intermediate event structure through node seeding, MEHG construction, and role binding.}
    \label{fig:intro}
\end{figure}

% \begin{figure}[t]
%     \centering
%     \includegraphics[width=1.0\linewidth]{figures/intro.pdf}
%     \caption{Argument Role F1 on the M2E2 benchmark~\cite{WASE} under Direct Prompting. Unless otherwise noted, all direct-prompting baselines use the same default vision tool, \texttt{Qwen3-VL-8B-Thinking}: text-only LLMs consume tool-generated visual context and box proposals, while LVLMs process the raw image together with the same auxiliary tool outputs. X-MTL~\cite{X-MTL} and ECHO are included for reference.}
%     \label{fig:intro}
% \end{figure}

\section{Introduction}
\label{sec:introduction}
Multimedia Event Extraction (M2E2) is the task of understanding what happens in a multimedia news report from both its text and accompanying image. 
Given a paired text-image input, M2E2 aims to identify the event trigger, determine the event type, and extract the participating arguments with role labels grounded in textual spans or visual regions~\cite{WASE}. 
Unlike mention-level tasks such as entity extraction \cite{yadav-bethard-2018-ner-survey,lample-etal-2016-ner}, which can often be solved through local span-level decisions, M2E2 must assemble textual and visual evidence into a coherent event structure~\cite{wadden2019dygiepp,lin2020jointglobalie}. 
By converting scattered multimodal evidence into structured event records, M2E2 can support downstream applications such as news understanding \cite{wang-etal-2022-n24news,muller-budack-etal-2021-news-analytics}, event retrieval \cite{voskarides-etal-2022-event-retrieval,gottschalk-demidova-2018-eventkg}, and crisis monitoring \cite{alam-ofli-imran-2018-crisismmd,abavisani-etal-2020-crisis-categorization}.

Despite recent progress, prevailing M2E2 paradigms still face two core limitations. 
{First}, as illustrated in Figure~\ref{fig:intro}(a), most methods formulate extraction mainly as cross-modal alignment problems, improving local mention--region matching but not explicitly maintaining event-level consistency~\cite{li2025multi,BARE,X-MTL}. 
Therefore, locally plausible evidence can still be assembled into an incoherent event hypothesis, and this structural error then propagates to event typing and argument-role prediction.

{Second}, LLM-based M2E2 approaches typically rely on direct prompting, which entangles event identification, cross-modal linking, and role binding within a single-pass generation process~\cite{yu2025m2e2ke,chen2025stepwise,wang2025mgfsg}. 
As illustrated in Figure~\ref{fig:intro}(b), this tightly coupled formulation can prematurely commit unsupported arguments to the final event record because early typing and grounding errors remain uninspected and unrevised. 
% 
% Furthermore, this limitation is paradigmatic that M2E2 requires structured inference over event hypotheses, whereas direct prompting casts it as one-shot generation. 
Furthermore, this limitation reflects a deeper paradigmatic mismatch that M2E2 requires structured inference over event hypotheses, whereas direct prompting casts it as one-shot generation.

To address this, iterative refinement over an explicit event structure is a more appropriate direction. 
For instance, most  approaches \cite{wu2024autogen,madaan2023selfrefine,shinn2023reflexion} based on Multi-Agent Systems (MAS) attempt to address these challenges through dialogue, debate, or role-playing. 
However, these approaches still rely on free-form natural-language state, whereas iterative refinement needs to maintain and update explicit event hypotheses under schema constraints \cite{liu2024lostmiddle}. 
Without a shared structured state, modifications proposed by different agents become difficult to align, verify, and roll back, while structurally critical decisions can be diluted as interaction history grows. 

Building on these observations, we present \textbf{ECHO}, a multi-agent framework for multimedia event extraction that centers on a shared \textbf{Multimedia Event Hypergraph (MEHG)}. 
Rather than leaving intermediate decisions in free-form dialogue, MEHG externalizes event hypotheses as explicit event-centric structures grounded in textual mentions and visual regions. This shared state gives all agents a common object of reasoning and revision. During inference, agents update the hypergraph through auditable atomic operations, including creating, revising, linking, unlinking, and pruning hypotheses. Each proposed update is checked against the current structure and the available multimodal evidence before it is committed. In this way, ECHO turns collaboration into controlled structural refinement and makes intermediate decisions explicit, verifiable, and reversible under schema constraints. 

ECHO further introduces a \textbf{Link-then-Bind} strategy to reduce cascading errors in multimodal extraction. 
The framework first stabilizes which textual mentions and visual regions belong to each event hypothesis, and only then assigns fine-grained semantic roles. 
This ordering is important because argument relevance and role identity do not become reliable at the same time under noisy cross-modal evidence. 
By separating relevance discovery from role commitment, ECHO can correct event identity and argument membership before semantic decisions become fixed. 
The result is a more stable extraction process that reduces premature role binding and limits the downstream impact of early grounding errors. 
Extensive experiments show that ECHO consistently outperforms existing baselines across textual, visual, and multimedia settings. 
% Its gains are especially pronounced for argument-role extraction, where it establishes new state-of-the-art results and improves Event Mention and Argument Role F1 by up to 7.3 and 15.5 points, respectively.

Our main contributions are as follows:
\begin{itemize}
    \item We introduce \textbf{MEHG} together with an operation-mediated multi-agent framework that performs auditable atomic updates over a shared hypergraph, making multimodal event hypotheses explicit, revisable, and verifiable.
    % \item 
    \item We propose \textbf{Link-then-Bind}, a deferred role-binding mechanism that separates event-argument relevance discovery from semantic role commitment, reducing premature binding and cascading extraction errors under uncertain cross-modal evidence.
    \item Extensive experiments on M2E2 show that ECHO consistently outperforms strong task-specific and generative baselines across textual, visual, and multimedia settings, with especially large gains on argument-role extraction.
\end{itemize}

\section{Related Work}

\subsection{Multimedia Event Extraction}
% M2E2 extracts event triggers/types and role-labeled arguments from paired text--image inputs.
Prior work largely follows two threads: strengthening cross-modal representations and improving learning signals.
Several methods align textual entities with visual objects by building shared semantic spaces, e.g., through reasoning over cross-modal graphs~\cite{WASE,MGIM} or contrastive alignment and grounding objectives~\cite{UniCL,Clip-event,BARE}.
Subsequent approaches focus on optimization and supervision, including cross-modal data augmentation~\cite{CAMEL} and multi-task learning for multimodal understanding, including joint modeling of M2E2 subtasks~\cite{X-MTL,11210197}.
In parallel, template-based formulations recast extraction as natural-language prediction~\cite{MMUTF}.

Recent works explore incorporating LLMs into M2E2 pipelines. Some approaches focus on parameter-level adaptations, such as editing LLM behaviors via knowledge editing~\cite{yu2025m2e2ke} or deploying collaborative multi-LoRA experts~\cite{yuan2025collaborative}. Concurrently, there is a growing trend of decomposing the complex extraction process into manageable stages, including stepwise schema-guided prompting frameworks~\cite{chen2025stepwise} and multi-grained scene graph enhancements~\cite{wang2025mgfsg}. 
Our work is complementary to these emerging paradigms. Instead of updating backbone parameters (e.g., via instruction tuning or knowledge editing) or relying on rigid sequential prompts, we study how training-free agentic collaboration can iteratively refine schema-constrained hypotheses over an explicit intermediate structure.

\subsection{Multi-Agent Systems}
LLM-based multi-agent systems are often coordinated through natural-language interaction, such as debate, role-playing, and self-refinement~\cite{park2023generative,liang-etal-2024-encouraging,chen2024reconcile,madaan2023selfrefine}. Beyond open-ended reasoning, agentic designs have also been applied to information extraction (IE) and structured prediction, including agents that update structured databases~\cite{jiao-etal-2024-text2db}, multi-agent deliberation for zero-shot relation triplet extraction~\cite{xu-etal-2024-two}, and debate-style coordination for zero-shot IE across types or tasks~\cite{lu-etal-2025-crossagentie}. More recent frameworks further move beyond pure chat histories by coordinating agents through external tools, editable artifacts, or shared working memory~\cite{wu2024autogen,hong2024metagpt}.

Related work in Event Extraction (EE) and Agentic Reasoning further suggests the value of graph-structured external state. In event extraction, graph-based formulations explicitly encode triggers, arguments, and their dependencies, including graph-parsing approaches to sentence-level EE~\cite{you-etal-2023-jseegraph}, heterogeneous or interaction graphs for document-level EE~\cite{xu-etal-2021-document,pan-etal-2024-document}, and hypergraph neural networks for higher-order structured IE~\cite{yan-etal-2023-joint}. In LLM-based reasoning and agents, knowledge- or memory-graph structures have likewise been used to organize intermediate reasoning and retrieval, such as knowledge-graph prompting~\cite{wen-etal-2024-mindmap}, editable memory graphs for personalized agents~\cite{wang-etal-2024-crafting}, and autonomous reasoning frameworks over knowledge graphs~\cite{jiang-etal-2025-kg}. We follow this structured-state line, but instantiate the shared artifact as a Multimedia Event Hypergraph (MEHG). Unlike prior graph memories or graph-based IE formulations, MEHG is designed for schema-constrained multimodal extraction and supports hypergraph edits that revise event hypotheses and cross-modal relevance before final role binding.

\begin{figure*}[t]
    \centering
    \includegraphics[width=1\linewidth]{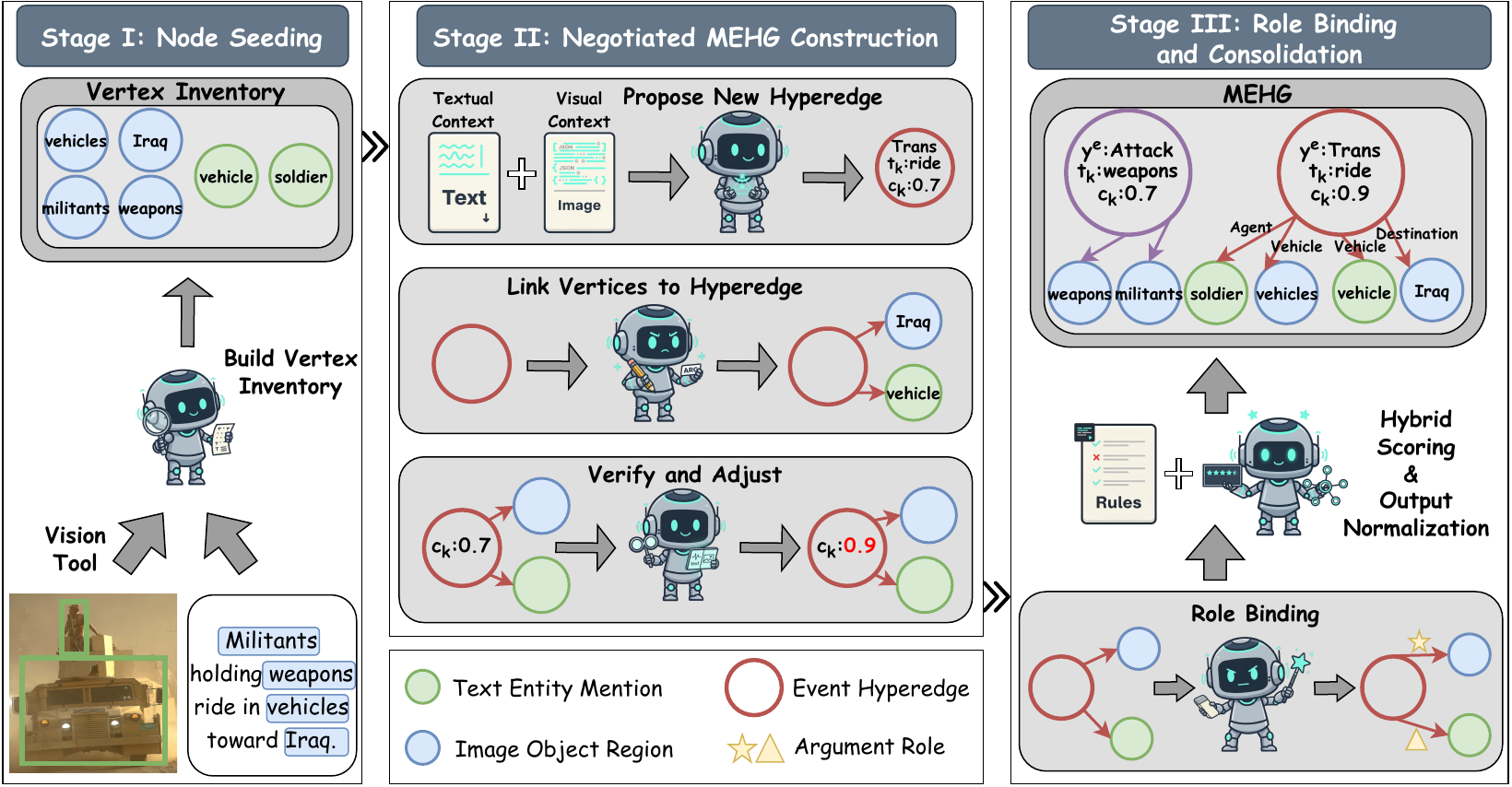}
    \caption{\textbf{Overview of ECHO.}
 Given $D=(T,I)$, Stage I constructs the vertex inventory and initializes an edge-free MEHG; Stage II agents negotiate MEHG updates via auditable atomic operations; Stage III performs role binding and consolidation to produce schema-consistent event predictions. The purple hyperedge represents a concurrent event hypothesis generated via the same ecosystem. Inside the hyperedge, $y^e$, $t_k$, and $c_k$ denote the event type, trigger, and confidence score, respectively.}
    \label{fig:ECHO}
\end{figure*}

\section{Prerequisite}
\label{sec:prerequisite}

\subsection{Task Definition}
\label{sec:task_definition}
Given a multimedia document $D=(T,I)$, where $T$ is the text and $I$ is the associated image, M2E2 aims to extract a set of structured event records from the paired input.
Each event record specifies a trigger span, an event type, and a set of grounded arguments with role labels.
An argument may be grounded either in a textual span in $T$ or a visual region in $I$, and its role must be valid under the schema of the predicted event type.
Formally, we denote the predicted event set by $\mathcal{S}=\{e_1,\ldots,e_m\}$, where each event $e_k=(t_k, y_k^e, A_k)$ consists of a trigger $t_k$, an event type $y_k^e \in \mathcal{Y}^e$, and a role-labeled argument set $A_k=\{(a_{k,j}, y_{k,j}^r)\}_{j=1}^{N_k}$ with $y_{k,j}^r \in \mathcal{Y}^r$.

% A valid prediction must satisfy both schema and grounding constraints: role assignments must be compatible with the event type, and each predicted argument must be supported by modality-specific evidence rather than free-form generation.
As emphasized in prior structured information extraction, correctness is therefore not an isolated local match, but whether the extracted pieces jointly form a coherent event record~\cite{wadden2019dygiepp,lin2020jointglobalie}.
In M2E2, this requirement is even stricter because triggers, role-labeled arguments, textual spans, and visual regions must be jointly resolved under a shared event schema across textual, visual, and multimedia settings.
The resulting prediction unit is thus naturally an event instance with a variable-size multimodal argument set, which calls for an explicit event-centric intermediate representation.

\subsection{Multimedia Event Hypergraph}
\label{sec:mehg}

Prior methods are mainly designed around cross-modal alignment and local association through graph-based interactions, contrastive objectives, or multi-task formulations~\cite{WASE,UniCL,Clip-event,X-MTL}.
However, the target of M2E2 is a structured event record rather than a set of independent local decisions, much like the event-level outputs studied in prior information extraction work~\cite{wadden2019dygiepp,lin2020jointglobalie}.
For example, one event may involve a single trigger together with several textual and visual arguments that must be interpreted under the same event schema, which is cumbersome to express directly with only linear outputs or collections of binary relations.
We therefore introduce a Multimedia Event Hypergraph (MEHG) as the shared intermediate representation for ECHO, so that each event hypothesis can be represented explicitly as a first-class object together with its currently relevant multimodal evidence.

Formally, given $D=(T,I)$, we construct an attributed hypergraph $H=(V,E)$, where $V$ is the set of grounded candidate vertices and $E$ is the set of event hyperedges; $E$ may be empty at initialization and is populated during negotiation.
The vertex set is defined as $V = V_T \cup V_I$, where $V_T$ contains textual candidate mentions and $V_I$ contains visual candidate regions.
Each vertex $v \in V$ is associated with a localization function $\ell(v)$, which maps textual vertices to extractive spans in $T$ and visual vertices to localized image regions.
Each hyperedge $\epsilon_k \in E$ represents a single event hypothesis and is defined as
\begin{equation}
\epsilon_k = (V_k, t_k, y_k^e, A_k, c_k),
\end{equation}
where $V_k \subseteq V$ is the set of currently linked candidate vertices, $t_k$ is the event trigger span, $y_k^e \in \mathcal{Y}^e$ is the event type, $c_k \in [0,1]$ is the confidence of the hypothesis, and $A_k$ records the committed role assignments for linked vertices:

\begin{equation}
A_k=\{(v, y_{k,v}^r)\mid v\in V_k,\; y_{k,v}^r \in \mathcal{Y}^r(y_k^e)\}.
\end{equation}

Here, $V_k$ tracks the candidates currently associated with an event hypothesis, while $A_k$ records the subset whose roles have been committed.
This separation allows the system to adjust event structure and candidate membership before finalizing role assignments.
MEHG thus serves as a shared, revisable workspace for intermediate event hypotheses rather than merely a final output representation. Compared with chain-structured representations, this event-centric state more naturally preserves a complete event hypothesis together with its multimodal arguments.

\section{Method}
\label{sec:method}

As illustrated in Figure~\ref{fig:ECHO}, ECHO is a training-free framework for M2E2 that iteratively refines event hypotheses over a shared Multimedia Event Hypergraph (MEHG). Given a multimedia document, the framework proceeds in three stages. Stage~I seeds textual and visual candidates, together with auxiliary visual context, to initialize an edge-free MEHG $H^{(0)}$. Stage~II refines this shared state into a stabilized MEHG $H^{(\star)}$, and Stage~III binds schema-compatible roles and consolidates the hypotheses into the final event set $\mathcal{S}$.

Most existing agentic frameworks coordinate through free-form dialogue or other implicit natural-language interaction~\cite{wu2024autogen,madaan2023selfrefine,shinn2023reflexion,yao2023tree}. For M2E2, this style of coordination, together with prediction schemes that couple relevance discovery and role assignment, can make early grounding errors hard to revise and easy to propagate into the final event structure~\cite{WASE,X-MTL}. ECHO instead uses an operation-mediated multi-agent protocol in which agents update a shared state through auditable atomic operations, making intermediate refinement explicit and easier to correct when needed. On top of this protocol, Link-then-Bind first stabilizes relevance between events and arguments and hypergraph topology before fine-grained role assignment, helping reduce the structural drift common in dialogue-mediated coordination. We will elaborate on this protocol in the following subsections.

\subsection{Stage I: Grounded Candidate Seeding}
 \label{sec:stage1}

Stage~I first performs multimodal candidate extraction on the input document to construct the initial shared state for later refinement. This stage takes the paired text and image as input, and outputs textual candidate vertices $V_T$, visual candidate vertices $V_I$, and auxiliary visual context $\mathcal{C}_I$.

On the text side, a seeding agent extracts candidate mentions from $T$ and instantiates textual vertices $V_T$, each grounded to an extractive surface span; repeated identical spans are deduplicated in the initial state. On the image side, an off-the-shelf vision tool proposes salient object regions as visual vertices $V_I$ and produces a compact natural language summary $\mathcal{C}_I$ for downstream reasoning. We then set $V^{(0)} = V_T \cup V_I$ and initialize the edge-free MEHG $H^{(0)}=(V^{(0)},E^{(0)})$ with $E^{(0)}=\emptyset$.

In Figure~\ref{fig:ECHO}, this stage creates textual vertices such as militants, weapons, vehicles, and Iraq, together with visual vertices such as vehicle. This stage is recall-oriented: it establishes a shared inventory of multimodal evidence while deferring event construction, relevance decisions, and role assignment to later stages.

\subsection{Stage II: Negotiated Hypergraph Construction}
\label{sec:phase2}

Stage~II starts from the initialized MEHG $H^{(0)}$ and iteratively refines event hypotheses by updating event topology and candidate relevance. Meanwhile, the fine-grained role commitment is strictly deferred to Stage~III. At coordination round $t$, the system maintains the current hypergraph state $H^{(t)}$ and an audit trail $\mathcal{L}^{(t)}$ that records previously committed updates. Conditioned on the document text $T$, the visual context $\mathcal{C}_I$, and the current state $(H^{(t)}, \mathcal{L}^{(t)})$, agents propose candidate updates to the MEHG as atomic operations.

\noindent \textbf{Atomic operation space.}
Stage~II is restricted to strictly topological edits: it may create or drop a hyperedge, revise its trigger span or event type, link or unlink a vertex to or from a hyperedge, and adjust the confidence of an existing hypothesis. For each event hyperedge $\epsilon_k=(V_k,t_k,y_k^e,\emptyset,c_k)$, Stage~II updates only $V_k$, $t_k$, $y_k^e$, and $c_k$; the role field remains unresolved. This is the essence of Link-then-Bind: the system can correct event identity and candidate membership without undoing prematurely assigned roles.

\noindent \textbf{Three-agent negotiation.}
Three agents operate on the same shared state from complementary perspectives. The Proposer introduces or revises event hypotheses, the Linker updates the relevance set $V_k$ for each hypothesis, and the Verifier cross-checks hypotheses against textual and visual evidence, calibrates confidence, and prunes weak or contradictory structures.

The middle panel of Figure~\ref{fig:ECHO} illustrates a typical round. The Proposer may introduce two competing hyperedges: a lower-confidence \texttt{Conflict:Attack} hypothesis and a stronger \texttt{Movement:Transport} hypothesis. The Linker then attaches vertices such as Iraq and vehicle to the transport hyperedge because they are directly supported by the text and image, while leaving their roles unspecified. The Verifier further raises the transport confidence. This occurs because the trigger, destination mention, and vehicle region are mutually supportive across modalities. Conversely, the attack hypothesis is kept at a lower confidence due to its weaker support. Because no role has yet been committed, either hypothesis can still be revised or removed in later rounds.

\noindent \textbf{Commit policy and audit trail.}
Let $\Delta^{(t)}$ denote the set of candidate operations proposed at round $t$. Before applying them, we perform lightweight structural checks to ensure that each operation is well-formed, references valid vertices or hyperedges, and satisfies basic schema constraints. We then resolve conflicts deterministically: \texttt{drop} overrides other modifications to the same hyperedge, \texttt{unlink} overrides \texttt{link} for the same vertex and hyperedge pairs, and conflicting \texttt{revise} operations defer to the Verifier. Let $\widetilde{\Delta}^{(t)}$ denote the surviving non-conflicting operation set. The committed update is
\begin{equation}
H^{(t+1)}=\mathrm{Apply}(H^{(t)},\widetilde{\Delta}^{(t)}), \qquad
\mathcal{L}^{(t+1)}=\mathcal{L}^{(t)} \cup \mathrm{Log}(\widetilde{\Delta}^{(t)}).
\end{equation}
Agents condition on $\mathcal{L}^{(t)}$ and are constrained not to repeat equivalent committed updates.

\noindent \textbf{Stopping criterion.}
Stage~II terminates when no valid operation survives checking and conflict resolution, i.e., $\widetilde{\Delta}^{(t)}=\emptyset$. We denote the resulting stabilized MEHG by $H^{(\star)}$.

\begin{table*}[t]
\centering
\caption{Main results of ECHO on event mention and argument role across modalities.
Best and second-best F1 are in \best{bold} and \second{underlined}, respectively. DeepSeek uses DeepSeek-V3.2.}
\label{tab:main_results}
{
\setlength{\tabcolsep}{3.5pt}
\begin{tabular}{@{} l c
    *{3}{r}*{3}{r}@{\hspace{0.6em}}
    *{3}{r}*{3}{r}@{\hspace{0.6em}}
    *{3}{r}*{3}{r} @{}}
\toprule
\multirow{3}{*}{\textbf{Model}} & \multirow{3}{*}{\textbf{Venue}} &
\multicolumn{6}{c}{\textbf{Textual Events}} &
\multicolumn{6}{c}{\textbf{Visual Events}} &
\multicolumn{6}{c}{\textbf{Multimedia Events}} \\
\cmidrule(lr){3-8} \cmidrule(lr){9-14} \cmidrule(lr){15-20}
& & \multicolumn{3}{c}{Event Mention} & \multicolumn{3}{c}{Argument Role}
& \multicolumn{3}{c}{Event Mention} & \multicolumn{3}{c}{Argument Role}
& \multicolumn{3}{c}{Event Mention} & \multicolumn{3}{c}{Argument Role} \\
\cmidrule(lr){3-5} \cmidrule(lr){6-8}
\cmidrule(lr){9-11} \cmidrule(lr){12-14}
\cmidrule(lr){15-17} \cmidrule(lr){18-20}
& & P & R & F1 & P & R & F1 & P & R & F1 & P & R & F1 & P & R & F1 & P & R & F1 \\
\midrule

Flat & ACL'20
& 34.2 & 63.2 & 44.4 & 20.1 & 27.1 & 23.1
& 27.1 & 57.3 & 36.7 & 4.3 & 8.9 & 5.8
& 33.9 & 59.8 & 42.2 & 12.9 & 17.6 & 14.9 \\

WASE-T & ACL'20
& 42.3 & 58.4 & 48.2 & 21.4 & 30.1 & 24.9
& \multicolumn{6}{c}{\NA}
& \multicolumn{6}{c}{\NA} \\

WASE-V\_att & ACL'20
& \multicolumn{6}{c}{\NA}
& 29.7 & 61.9 & 40.1 & 9.1 & 10.2 & 9.6
& \multicolumn{6}{c}{\NA} \\

WASE-V\_obj & ACL'20
& \multicolumn{6}{c}{\NA}
& 28.6 & 59.2 & 38.7 & 13.3 & 9.8 & 11.2
& \multicolumn{6}{c}{\NA} \\

WASE-att & ACL'20
& 37.6 & 66.8 & 48.1 & 27.5 & 33.2 & 30.1
& 32.3 & 63.4 & 42.8 & 9.7 & 11.1 & 10.3
& 38.2 & 67.1 & 49.1 & 18.6 & 21.6 & 19.9 \\

WASE-obj & ACL'20
& 42.8 & 61.9 & 50.6 & 23.5 & 30.3 & 26.4
& 43.1 & 59.2 & 49.9 & 14.5 & 10.1 & 11.9
& 43.0 & 62.1 & 50.8 & 19.5 & 18.9 & 19.2 \\

CLIP-Event & CVPR'22
& \multicolumn{6}{c}{\NA}
& 41.3 & 72.8 & 52.7 & 21.1 & 13.1 & 17.1
& \multicolumn{6}{c}{\NA} \\

UniCL & MM'22
& 49.1 & 59.2 & 53.7 & 27.8 & 34.3 & 30.7
& 54.6 & 60.9 & 57.6 & 16.9 & 13.8 & 15.2
& 44.1 & 67.7 & 53.4 & 24.3 & 22.6 & 23.4 \\

MGIM & TCSVT'24
& 50.1 & 66.5 & 55.8 & 28.2 & 34.7 & 31.2
& 55.7 & 64.4 & 58.5 & 24.1 & 14.1 & 17.8
& 46.3 & 69.6 & 55.6 & 25.2 & 21.7 & 24.6 \\

UMIE & AAAI'24
& \multicolumn{6}{c}{\NA}
& \multicolumn{6}{c}{\NA}
& \NA & \NA & 62.1 & \NA & \NA & 24.5 \\

MMUTF & EMNLP'24
& 48.5 & 65.0 & 55.5 & 33.6 & 44.2 & 38.2
& 55.1 & 59.1 & 57.0 & 23.6 & 18.8 & 20.9
& 47.9 & 63.4 & 54.6 & 39.9 & 20.8 & 27.4 \\

CAMEL & MM'23
& 45.1 & 71.8 & 55.4 & 24.8 & 41.8 & 31.1
& 52.1 & 66.8 & 58.5 & 21.4 & 28.4 & 24.4
& 55.6 & 59.5 & 57.5 & 31.4 & 35.1 & 33.2 \\

SSGPF & ICME'25
& \multicolumn{6}{c}{\NA}
& \multicolumn{6}{c}{\NA}
& 60.4 & 72.1 & 65.7 & 33.8 & 38.5 & 36.0 \\

X-MTL & AAAI'25
& 49.7 & 65.7 & 56.6 & 34.6 & 37.6 & 36.0
& 73.1 & 70.3 & 71.7 & 33.2 & 31.3 & 32.2
& 78.3 & 57.3 & 66.2 & 40.3 & 42.6 & 41.4 \\

\midrule
    \multicolumn{20}{@{}l}{\textbf{ECHO}} \\
-Qwen3-8B & -
& 62.3 & 64.3 & 63.3 & 32.2 & 53.8 & 40.3
& 78.6 & 78.2 & 78.4 & 61.6 & 60.0 & \best{60.8}
& 70.6 & 69.9 & 70.2 & 47.0 & 58.9 & 52.3 \\
-Qwen3-14B & -
& 63.7 & 66.1 & \second{64.9} & 33.6 & 53.1 & 41.1
& 77.7 & 77.7 & 77.7 & 61.9 & 59.6 & \second{60.7}
& 73.5 & 73.5 & 73.5 & 48.6 & 59.0 & 53.3 \\
-Qwen3-32B & -
& 60.9 & 66.1 & 63.4 & 32.7 & 56.2 & \best{41.4}
& 80.3 & 80.3 & \second{80.3} & 61.8 & 57.6 & 59.6
& 72.5 & 72.5 & 72.5 & 49.8 & 61.2 & \second{54.9} \\
-GPT-5 & -
& 61.6 & 71.0 & \best{66.0} & 27.2 & 62.3 & 37.9
& 82.4 & 81.9 & \best{82.1} & 61.8 & 58.3 & 60.0
& 79.6 & 79.6 & \best{79.6} & 44.8 & 63.5 & 52.5 \\
-DeepSeek & -
& 62.0 & 67.5 & 64.6 & 32.3 & 57.2 & \second{41.3}
& 82.4 & 81.9 & \best{82.1} & 62.7 & 58.1 & 60.3
& 75.1 & 75.1 & \second{75.1} & 47.2 & 65.8 & \best{55.0} \\
\bottomrule
\end{tabular}
}
\end{table*}

\subsection{Stage III: Role Binding and Consolidation}
\label{sec:stage3}

Stage~III takes the stabilized MEHG $H^{(\star)}$ and completes each event hypothesis by instantiating its role field, followed by final scoring and output normalization.

\noindent \textbf{Role binding.}
For each stabilized hyperedge $\epsilon_k=(V_k,t_k,y_k^e,\emptyset,c_k)\\\in H^{(\star)}$, we predict an event-type-specific role $y_{k,v}^r \in \mathcal{Y}^r(y_k^e)\cup\{\varnothing\}$ and an argument-level confidence $s_{k,v}\in[0,1]$ for each linked vertex $v\in V_k$. For textual vertices, role prediction is conditioned on $(T,t_k,y_k^e)$, with $\mathcal{C}_I$ used as auxiliary context when needed. For visual vertices, we query the vision tool with the event hypothesis, then align the returned localized candidates to linked visual vertices by localization consistency (e.g., overlap), discarding unmatched proposals. We retain only non-empty bindings with $s_{k,v}\ge\tau$, yielding $A_k=\{(v,y_{k,v}^r)\mid v\in V_k,\; y_{k,v}^r\neq\varnothing,\; s_{k,v}\ge\tau\}$. After binding, we denote the completed hypothesis by $\hat{\epsilon}_k=(V_k,t_k,y_k^e,A_k,c_k)$.

In Figure~\ref{fig:ECHO}, the stabilized transport hypothesis yields militants (\texttt{Agent}), vehicles (\texttt{Vehicle}), and Iraq (\texttt{Destination}); weapons receives $\varnothing$ and is excluded from $A_k$.

\noindent \textbf{Final scoring and consolidation.}
We assign each completed hypothesis $\hat{\epsilon}_k$ a final score $c_k^{\mathrm{final}}\in[0,1]$ by combining its negotiated confidence from Stage~II, the retained argument level evidence, and a lightweight schema-consistency term:

\begin{equation}
\label{eq:hybrid_score}
c_k^{\mathrm{final}}
=
\alpha\, c_k
+
(1-\alpha-\lambda)\,\mathrm{AggConf}(A_k)
+
\lambda\,\mathrm{RuleScore}(\epsilon_k, H^{(\star)}),
\end{equation}
% where $\mathrm{AggConf}(A_k)$ is the average retained argument confidence in $A_k$ (or $0$ if $A_k=\emptyset$), and $\mathrm{RuleScore}(\epsilon_k, H^{(\star)})\in[0,1]$ is a normalized heuristic over schema validity and support consistency. After scoring all candidate hyperedges for an instance, we retain $\epsilon_k$ only if $c_{\max}^{\mathrm{final}}-c_k^{\mathrm{final}}\le 0.15$, where $c_{\max}^{\mathrm{final}}$ is the highest final score among the candidates. Unless otherwise noted, we use $\alpha=0.5$, $\lambda=0.1$, and $\tau=0.7$.
where $\mathrm{AggConf}(A_k)$ is the average retained argument confidence in $A_k$ (or $0$ if $A_k=\emptyset$), and $\mathrm{RuleScore}(\hat{\epsilon}_k)\in[0,1]$ is a normalized heuristic favoring schema-valid, mutually supported bindings. To avoid suppressing distinct events in the same document, consolidation is restricted to a local competition set $\mathcal{C}(k)$ consisting of hyperedges with the same trigger span or substantially overlapping evidence. We keep $\hat{\epsilon}_k$ when $\max_{\hat{\epsilon}_j\in \mathcal{C}(k)} c_j^{\mathrm{final}} - c_k^{\mathrm{final}} \le \delta$, with $\delta=0.15$. In Figure~\ref{fig:ECHO}, this retains the better supported transport hypothesis over the weaker attack alternative. Unless otherwise noted, we set $\alpha=0.5$, $\lambda=0.1$, and $\tau=0.7$.

\noindent \textbf{Output normalization.}
For evaluation, we align triggers and textual arguments to minimal surface spans in $T$ to reduce paraphrasing and over-extended mentions. This extractive cleanup is performed after role binding, so it does not perturb the event topology. Each retained hypothesis is then exported as a final event record $e_k=(t_k,y_k^e,A_k)$, optionally with confidence scores derived from $c_k^{\mathrm{final}}$ and $\{s_{k,v}\}$.

\section{Experiments}
\label{sec:experiments}

We first describe the experimental setup, including the dataset, evaluation protocol, baselines, and controlled comparison settings, and then answer the following research questions:
\begin{itemize}\setlength\itemsep{0pt}
    \item \textbf{RQ1}: Does ECHO achieve state-of-the-art performance on M2E2 across different backbones?
    \item \textbf{RQ2}: Under the same grounding and output constraints, how do Direct Prompting and dialogue-mediated multi-agent coordination compare with ECHO?
    \item \textbf{RQ3}: Which components are critical to ECHO, especially the Link-then-Bind schedule?
    \item \textbf{RQ4}: How efficient is iterative hypergraph construction in terms of convergence and cost?
\end{itemize}

\begin{figure*}[t]
    \centering
    \includegraphics[width=1.0\linewidth]{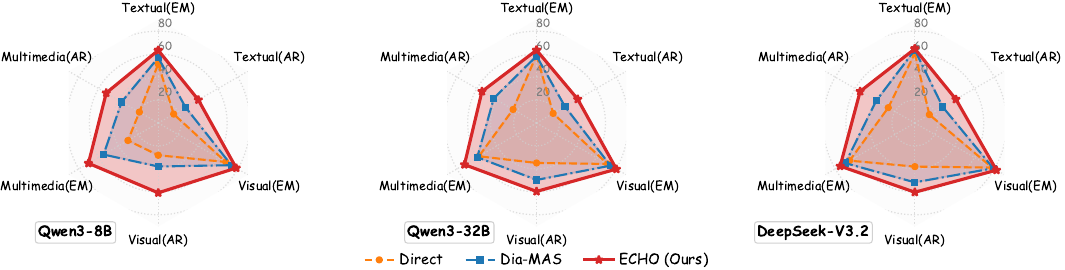}
    \caption{F1 comparison of Direct prompting, Dialogue-Mediated baseline, and ECHO on M2E2 across all settings.}
    \label{fig:empowering_effect}
\end{figure*}

\subsection{Experimental Settings}
\label{sec:exp-settings}

\noindent\textbf{Dataset and evaluation.}
We evaluate on the M2E2 benchmark~\cite{WASE}. Each example is a multimedia news document $D=(T,I)$ annotated with 8 event types and 15 argument roles. The dataset contains 245 documents, including 1,297 textual events, 391 visual events, and 309 coreferenced multimedia events. Following the official evaluation script, we report Precision (P), Recall (R), and F1 on (i) Event Mention (EM) and (ii) Argument Role (AR) under textual, visual, and multimedia settings. A visual argument is counted as correct if its predicted region overlaps the gold box with IoU $\ge 0.5$.

\noindent\textbf{Baselines.}
We compare ECHO with two groups of baselines. The first group comprises task-specific M2E2 systems: Flat, the WASE family~\cite{WASE}, UniCL~\cite{UniCL}, CLIP-Event~\cite{Clip-event}, CAMEL~\cite{CAMEL}, MMUTF~\cite{MMUTF}, UMIE~\cite{Umie}, MGIM~\cite{MGIM}, and X-MTL~\cite{X-MTL}. We also include two recent LLM-based M2E2 methods, SSGPF~\cite{chen2025stepwise} and LLM Knowledge Editing~\cite{yu2025m2e2ke}; the latter is discussed separately in Appendix~B.1 because its argument extraction stage uses gold event types. The second group consists of training-free generative baselines: Direct Prompting and a MetaGPT-style dialogue-mediated multi-agent baseline~\cite{hong2024metagpt}.

\noindent\textbf{Controlled comparison setup.}
Unless otherwise noted, all generative methods---ECHO, Direct Prompting, and the dialogue-mediated baseline---use the same default vision tool, \texttt{Qwen3-VL-8B-Thinking}, for visual context generation and localization. Text-only LLMs receive the document text together with the tool-generated image description and box proposals, while LVLMs receive the raw image together with the same auxiliary tool outputs. To ensure output-side fairness, we also enforce the same schema-constrained JSON contract across all generative settings. Therefore, the main difference in the controlled comparison is not access to visual evidence or output flexibility, but whether the method can iteratively revise an explicit structured intermediate state.

\noindent\textbf{Dialogue-mediated baseline.}
Our dialogue-mediated baseline is designed to match ECHO in backbones, multimodal inputs, and agent specialization. It uses the same seeded candidate inventory and evidence sources, but replaces atomic MEHG operations with natural-language dialogue over a shared textual working memory.

\noindent\textbf{Implementation details.}
Unless otherwise noted, we set $T_{\max}=2$ negotiation rounds with early stopping when no further valid operations are committed. All LLM/LVLM calls use the same decoding setup unless explicitly varied elsewhere. For completeness, runtime configuration is detailed in Appendix~A.2, controlled baseline and output-format details in Appendices~A.3--A.5, and extended experimental results in Appendices~B--D.

\subsection{Overall Performance Comparison (RQ1)}
\label{sec:rq1}

Table~\ref{tab:main_results} reports the main results on M2E2 under textual, visual, and multimedia settings.  Under comparable end-to-end settings, ECHO consistently outperforms prior task-specific systems across settings. Even with a moderate-sized backbone (Qwen3-8B), ECHO already surpasses X-MTL on all reported subtasks, with the largest gains on argument role extraction: Visual AR F1 nearly doubling to achieve a score of 60.8, and Multimedia AR F1 realizing an absolute improvement of 10.9 points to arrive at an F1 of 52.3.

ECHO remains effective across backbones. Scaling to Qwen3-32B yields further gains, and proprietary backbones further extend the upper bound: DeepSeek-V3.2 achieves the best Multimedia AR F1 (55.0), while GPT-5 delivers the strongest event mention performance in several settings. For stronger backbones, AR gains can be less pronounced because higher recall is sometimes accompanied by lower precision, indicating over-generation of role bindings. Detailed output statistics are reported in Appendix~C.3, and the error breakdown is provided in Appendix~C.2.

Compared with SSGPF~\cite{chen2025stepwise}, ECHO with a comparably sized Qwen3-8B backbone improves Event Mention F1 by 4.2 points and Argument Role F1 by 16.3 points on the comparable setting. We discuss LLM Knowledge Editing~\cite{yu2025m2e2ke} separately in Appendix~B.1, since its argument extraction stage uses gold event types and is therefore not strictly comparable in an end-to-end setting.

\subsection{Direct Prompting LLM/LVLM and Dialogue-Mediated Baselines (RQ2)}
\label{sec:rq2}
We next compare ECHO with Direct Prompting and the dialogue-mediated baseline under the controlled setting in Sec.~\ref{sec:exp-settings}. Figure~\ref{fig:empowering_effect} shows the representative comparison on the same backbones, while the full Direct Prompting and dialogue-mediated results for additional LLMs/LVLMs are reported in Appendices~B.2 and~B.3, respectively. Across both LLM and LVLM instantiations, Direct Prompting remains clearly weaker on argument role extraction, especially in visual and multimedia settings, suggesting that single-pass generation is brittle under strict schema and grounding constraints. In some cases, its event mention performance can be competitive, but the gap becomes much larger on argument role extraction, where grounded and schema-consistent role assignment is the main difficulty.

Figure~\ref{fig:empowering_effect} compares Direct Prompting, the dialogue-mediated baseline, and ECHO under the same backbones. The dialogue-mediated baseline often improves over Direct Prompting, suggesting that iterative interaction can partially correct early mistakes. However, ECHO still brings consistent gains, especially on argument role extraction. This suggests that the main advantage comes not only from multi-agent interaction itself, but from explicit state-mediated revision through MEHG operations.

\begin{figure}[t]
    \centering
    \includegraphics[width=1.0\linewidth]{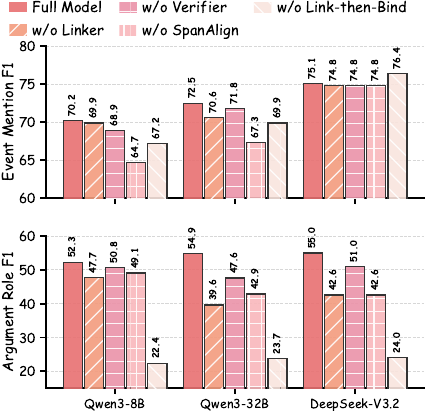}
    \caption{Ablation on the multimedia setting of M2E2. We report F1 for EM and AR with three backbones.}
    \label{fig:ablation_f1}
\end{figure}

\subsection{Ablation Study (RQ3)}
\label{sec:rq3}
We conduct ablations in the multimedia setting to assess the contribution of ECHO's key design choices, including Stage~II relevance negotiation, hypothesis verification, and Stage~III span normalization. Figure~\ref{fig:ablation_f1} summarizes EM/AR F1 across three backbones, while the full precision--recall--F1 breakdown is provided in Appendix~B.4.

\noindent\textbf{Link-then-Bind.}
We quantify the effect of Link-then-Bind by forcing the Linker to consider roles during linking, rather than deferring role commitment to Stage~III. Across all three backbones, this change substantially lowers AR F1 in the multimedia setting. The predicted argument set also becomes smaller (lower Pred/Gold), indicating that early role conditioning makes Stage~II overly conservative before the relevance topology has stabilized. We do observe a slight EM gain for DeepSeek-V3.2, suggesting that role cues can sometimes help consolidate event-level predictions. However, this benefit is outweighed by the larger loss in argument recall and final argument role accuracy. Overall, the results support our design choice of first stabilizing event structure and candidate relevance, and only then committing fine-grained semantic roles.

\noindent\textbf{Other components.}
Removing the Linker (replaced by all-to-all linking) further degrades AR, while EM is comparatively stable. This suggests that negotiated relevance primarily improves the quality of the candidate set passed to downstream role binding, rather than event detection itself. The Verifier also provides consistent gains, especially for stronger backbones that generate more diverse hypotheses; without explicit pruning, weak, redundant, or contradictory structures tend to persist and disproportionately hurt argument-role prediction. Finally, removing SpanAlign reduces extraction quality under the extractive M2E2 protocol, where span mismatches and over-extended mentions are directly penalized.

\noindent\textbf{Vision tool ablation.}
We further test whether ECHO's gains mainly come from the vision tool itself. Keeping the reasoning backbone fixed, we replace the default vision tool with a stronger alternative and report the resulting AR performance in Table~\ref{tab:vision_tool_ablation}. A stronger vision tool consistently improves AR in both visual and multimedia settings, confirming that better visual grounding is beneficial. However, this effect is complementary rather than explanatory: under the main controlled comparison setup, all generative methods use the same default vision tool, so ECHO's gains there cannot be attributed to privileged visual preprocessing alone.

\begin{table}[t]
\centering
\caption{Vision tool ablation on AR. Replacing the default vision tool with a stronger one consistently improves visual and multimedia AR.}
\label{tab:vision_tool_ablation}
\begin{tabular}{@{}lcccccc@{}}
\toprule
& \multicolumn{3}{c}{\shortstack[c]{Qwen3-VL-8B\\Thinking}} 
& \multicolumn{3}{c}{\shortstack[c]{Qwen3-VL-30B-A3B\\Thinking}} \\
\cmidrule(lr){2-4} \cmidrule(lr){5-7}
Setting & P & R & F1 & P & R & F1 \\
\midrule
\multicolumn{7}{c}{\textbf{Qwen3-8B}} \\
Visual      & 61.6 & 60.0 & 60.8 & 63.1 & 62.5 & 62.8 (+2.0) \\
Multimedia  & 47.0 & 58.9 & 52.3 & 49.5 & 64.5 & 56.0 (+3.7) \\
\midrule
\multicolumn{7}{c}{\textbf{Qwen3-32B}} \\
Visual      & 61.8 & 57.6 & 59.6 & 66.0 & 60.0 & 62.9 (+3.3) \\
Multimedia  & 49.8 & 61.2 & 54.9 & 52.3 & 62.7 & 57.0 (+2.1) \\
\bottomrule
\end{tabular}
\end{table}

\subsection{Budget and Efficiency Analysis (RQ4)}
\label{sec:rq4}

\begin{figure}[t]
  \centering
  \includegraphics[width=1.0\linewidth]{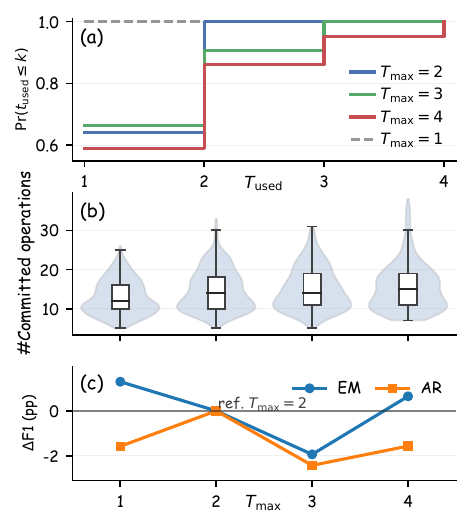}
  \caption{
  Negotiation budget analysis for Stage~II under early stopping.
  (a) Proportion of samples that stop by round $k$ (i.e., $T_{\mathrm{used}} \le k$) when varying the maximum budget $T_{\max}$.
  (b) Distribution of the number of committed atomic operations per sample.
  (c) Sensitivity of extraction performance to $T_{\max}$, reported as $\Delta$F1 (percentage points) relative to $T_{\max}{=}2$.
  }
  \label{fig:negotiation_budget}
\end{figure}

RQ4 examines whether Stage~II negotiation remains controlled in practice and what efficiency cost it introduces. Figure~\ref{fig:negotiation_budget} analyzes three aspects: (i) the empirical stopping distribution over used rounds $T_{\text{used}}$, (ii) the number of committed atomic operations per sample, and (iii) sensitivity to the maximum round budget $T_{\max}$ under early stopping.

Figure~\ref{fig:negotiation_budget}a shows that most samples stop within a small number of rounds under the criterion $\Delta^{(t)}=\emptyset$, indicating that the MEHG typically stabilizes quickly in practice. Figure~\ref{fig:negotiation_budget}b further shows that increasing $T_{\max}$ mainly affects a relatively small tail of harder cases, while the median number of committed operations grows only moderately. This suggests that iterative refinement remains bounded under a limited negotiation budget rather than becoming open-ended. Figure~\ref{fig:negotiation_budget}c shows that performance is relatively insensitive to $T_{\max}$ in the 1--4 range, supporting $T_{\max}=2$ as a practical default that balances quality and efficiency.

To complement this analysis, we report cost profiles in Appendix~D.1, including interaction cost (main LLM and vision tool calls) and computation cost (token usage) for ECHO and the dialogue-mediated baseline under the same settings. As expected, ECHO is less efficient than one-shot Direct Prompting by design, since it performs iterative negotiation and a separate role-binding stage rather than a single-pass generation step. However, this additional cost accompanies consistent gains in extraction quality, especially on argument role extraction. Appendix~D.1 further shows that ECHO is more efficient in token usage than the dialogue-mediated baseline under the same settings. Overall, these results suggest that Stage~II provides iterative structural refinement with bounded and manageable overhead.

\begin{table}[t]
\centering
\caption{AR error counts for Qwen3-32B under ECHO on M2E2.}
\label{tab:ar_error_qwen32b}
\begin{tabular}{lccc}
\toprule
\textbf{AR error type} & \textbf{Textual} & \textbf{Visual} & \textbf{Multimedia} \\
\midrule
Spurious                         & 540 & 0   & 180 \\
Span mismatch                    & 552 & 0   & 181 \\
Role misassignment               & 60  & 12  & 54  \\
No-gold event type               & 768 & 111   & 208  \\
Localization error  & 0   & 173 & 400 \\
\midrule
\textbf{Total}                   & 1920 & 296 & 1023 \\
\bottomrule
\end{tabular}
\end{table}

\subsection{Error Analysis}
\label{sec:error_analysis}

Table~\ref{tab:ar_error_qwen32b} summarizes AR error counts of ECHO with Qwen3-32B in the textual, visual, and multimedia settings. We consider five error categories: (1) \textit{Spurious}, where the system predicts extra arguments unsupported by any gold role; (2) \textit{Span mismatch}, where the predicted textual content is plausible but the extracted span does not match the gold field after normalization; (3) \textit{Role misassignment}, where the argument content is correct but assigned to an incorrect role; (4) \textit{No-gold event type}, where the system predicts arguments under event types with no corresponding gold arguments; and (5) \textit{Localization error}, where a predicted visual argument fails the scorer's box-matching criterion (IoU $\geq 0.5$).

The error distribution reveals different bottlenecks across settings. In textual AR, the dominant errors are \textit{No-gold event type}, \textit{Span mismatch}, and \textit{Spurious}, suggesting that the main failures come from over-generation and imperfect extractive realization rather than frequent role confusion. In visual and multimedia AR, \textit{Localization error} becomes the largest error source, indicating that visual grounding remains the main bottleneck once a plausible event structure has been formed.

The prevalence of \textit{Spurious} and \textit{No-gold event type} errors is also consistent with ECHO's design. Stages~I--II maintain a revisable high-recall candidate set, while the Verifier removes weak or contradictory hypotheses without aggressively pruning every plausible candidate. As a result, over-generation is reduced but not fully eliminated, and the remaining errors mainly affect precision. \textit{Span mismatch} is partially alleviated by the normalization step in Stage~III, whereas the residual \textit{Localization error} in vision-involved settings suggests that visual grounding is the main direction for further improvement.

\section{Conclusion}

We presented ECHO, a training-free multi-agent framework for M2E2 that incrementally constructs and revises a shared MEHG through atomic operations under schema and grounding constraints. ECHO follows a Link-then-Bind strategy: it first stabilizes event hypotheses and cross-modal relevance links, and then performs fine-grained role binding and global consolidation over the resulting structure. Across textual, visual, and multimedia evaluations on M2E2, ECHO achieves consistent gains over strong LLM and LVLM baselines. These results suggest that making intermediate structures explicit and deferring fine-grained commitment until the event topology is sufficiently stable is an effective way to reduce cascading errors in direct generative extraction. More broadly, our findings highlight the value of state-based multi-agent orchestration for multimodal structured extraction. In future work, we will extend MEHG-based reasoning to broader structured extraction tasks and further improve inference efficiency under bounded computation budgets.

\bibliographystyle{ACM-Reference-Format}
\bibliography{my_references}

\clearpage
\appendix
\section{Reproducibility and Implementation Details}
\label{sec:appendix-repro}

\subsection{Dataset and Evaluation Protocol (M2E2)}
\label{subsec:appendix-m2e2}

\noindent \textbf{Dataset.}
We evaluate on the M2E2 benchmark, where each instance is a news document paired with a single image.
The dataset uses a fixed event schema with 8 event types and 15 argument roles, and contains 245 documents in total, including 1,297 textual events, 391 visual events, and 309 coreferenced multimedia events.

\noindent \textbf{Event schema.}
Table~\ref{tab:m2e2_schema} summarizes the event types and their role inventories.

\begin{table}[h]
\centering
\caption{Event types and corresponding argument roles in M2E2.}
\label{tab:m2e2_schema}
\setlength{\tabcolsep}{6pt}
\renewcommand{\arraystretch}{1.25}
\begin{tabular}{|>{\raggedright\arraybackslash}p{0.42\columnwidth}
                |>{\raggedright\arraybackslash}p{0.45\columnwidth}|}
\hline
\textbf{Event Type} & \textbf{Argument Role} \\
\hline
Movement:Transport & Agent, Artifact, Vehicle, Destination, Origin \\
\hline
Conflict:Attack & Attacker, Target, Instru-\newline ment, Place \\
\hline
Conflict:Demonstrate & Entity, Police, Instru-\newline ment, Place \\
\hline
Justice:Arrest-Jail & Agent, Person, Instru-\newline ment, Place \\
\hline
Contact:PhoneWrite & Entity, Instrument, Place \\
\hline
Contact:Meet & Participant, Place \\
\hline
Life:Die & Agent, Instrument, Victim,\newline Place \\
\hline
Transaction:Transfer-\newline Money & Giver, Recipient, Money \\
\hline
\end{tabular}
\end{table}

\noindent \textbf{Evaluation settings.}
Following prior work, we report results under three settings:
(i) \emph{textual} (events grounded in text),
(ii) \emph{visual} (events grounded in the image), and
(iii) \emph{multimedia} (cross-modal/coreferenced events).
In the \emph{visual} setting, event mentions are evaluated without textual triggers; we predict the event type and grounded visual arguments. For clarity, the \emph{multimedia} setting evaluates a single cross-modal event record in which the textual and visual arguments are assembled as evidence for the same coreferenced event instance, rather than scored as two independent unimodal predictions. 

\noindent \textbf{Metrics.}
We follow the official M2E2 evaluation script and report Precision (P), Recall (R), and F1 for
(i) \emph{Event Mention} and (ii) \emph{Argument Role}.
An event is correct if it matches the gold event type and the required mention fields under the corresponding setting.
An argument is correct if the role label matches and the argument grounding matches the gold annotation (text span for textual arguments; bounding box for visual arguments).
For visual arguments, a predicted box is considered correct if its IoU with the gold box is at least 0.5. For textual fields, matching is performed on extractive spans after a final normalization pass that maps triggers and text arguments to minimal contiguous surface substrings in $T$; this pass does not alter event type, argument role, or event-argument membership. Controlled fairness details for generative methods (shared vision tool, unified JSON contract, and modality-integrity constraints) are provided later in this appendix.

\subsection{Runtime Configuration}
\label{app:runtime}

\noindent \textbf{Controlled comparison setup.}
All controlled comparisons among ECHO, Direct Prompting, and the dialogue-mediated baseline use the same default vision tool, Qwen3-VL-8B-Thinking, the same schema-constrained JSON output contract, and the same modality-integrity rules.
Separate tool-ablation results are reported only when the vision tool is explicitly varied.

\noindent \textbf{Decoding and budget.}
Unless otherwise specified, all LLM/LVLM calls use a temperature of $0.2$ with \texttt{max\_tokens}=8192.
All controlled comparisons are training-free: we do not apply task-specific fine-tuning, instruction tuning, parameter updates, or backbone-specific calibration to ECHO, Direct Prompting, or the dialogue-mediated baseline. 
For ECHO, Stage~II uses a maximum negotiation budget of $T_{\max}=2$ rounds, with early stopping when no non-conflicting valid operation survives the commit checks.
Unless otherwise stated, the final-scoring parameters in Eq.~(4) are fixed to $\alpha=0.5$, $\lambda=0.1$, and $\tau=0.7$ for all backbones and settings.

\subsection{Dialogue-Mediated Baseline}
\label{app:dialogue_baseline}

The dialogue-mediated baseline is inspired by dialogue-centric multi-agent coordination paradigms such as MetaGPT~\cite{hong2024metagpt}, but is instantiated here under the same controlled M2E2 setting as ECHO.
It is intentionally matched to ECHO in backbone family, multimodal evidence, seeded candidate inventory, agent specialization, and output contract; the only deliberate difference is that intermediate refinement is carried out through free-form dialogue over a shared textual working memory rather than through auditable MEHG operations.

\noindent \textbf{Stage I: Node Seeding.}
We use the exact same seeding prompt and produce the same types of initial candidates (text mentions, visual regions, and textual visual contexts) as inputs for the downstream steps.

\noindent \textbf{Stage II: Negotiation.}
We retain the same three specialized roles (Proposer, Linker, and Verifier) and provide them with the same multimodal evidence.
However, instead of proposing constrained atomic MEHG operations, agents communicate in free-form natural language and update a shared textual working memory that accumulates the current event hypotheses and linked evidence.
Unlike ECHO, this baseline does not maintain an explicit topological state whose event identity, candidate membership, and role assignment can be revised separately.

\noindent \textbf{Stage III: Scoring and output.}
The dialogue-mediated baseline does not enforce the Link-then-Bind schedule with a dedicated role-binder agent.
Instead, a single consolidator scores the dialogue-derived candidate hypotheses and emits the final JSON prediction under the unified output contract (Appendix~\ref{app:schema}).

\noindent \textbf{Cost profile note.}
Under the same round budget, the dialogue-mediated baseline uses one fewer main-model stage than ECHO because it replaces ECHO's dedicated Stage~III role-binding step with a single final consolidator.
However, it often consumes more tokens in practice, since intermediate state is carried implicitly in accumulated dialogue rather than compactly represented as MEHG edits.
For reference, Direct Prompting is cheaper still: after input preparation, it uses a single extraction call, which in the visual and multimedia settings corresponds to one vision-tool call plus one main-model call.

\subsection{Unified Task Format and Output Schema}
\label{app:schema}

\noindent \textbf{Evaluation-side output contract.}
To ensure evaluation-side fairness, we enforce the same schema-constrained JSON output contract across Direct Prompting, the dialogue-mediated baseline, and ECHO.
This contract applies only to the final prediction passed to the scorer.
In particular, ECHO still performs inference over an explicit MEHG with richer internal metadata, but all retained hypotheses are exported to the same normalized event-record format before evaluation.

Each prediction is formatted as a single JSON object containing an \texttt{events} list.
Each event must include an \texttt{event\_type}, a \texttt{trigger}, and two modality-separated argument lists: \texttt{text\_arguments} and \texttt{image\_arguments}.

\noindent \textbf{Modality integrity.}
Arguments must remain in their modality-specific fields throughout generation and normalization: text arguments are recorded only in \texttt{text\_arguments}, image arguments only in \texttt{image\_arguments}, and the two are never merged across modalities.

\noindent \textbf{Trigger and grounding constraints.}
When textual evidence is available, the \texttt{trigger} must be copied verbatim from $T$ as a contiguous surface substring; in the controlled generative setting, we further constrain it to a single token for all compared methods.
For image-only inputs, the \texttt{trigger} may be left empty or populated with a short visual descriptor.
In the visual setting, this field is retained only for a uniform output interface, since visual event mention scoring does not rely on textual triggers.
For visual arguments, every retained \texttt{image\_arguments} entry must include a grounded bounding box formatted as integers in \texttt{[x\_min, y\_min, x\_max, y\_max]} order.

\noindent \textbf{Scored fields vs.\ auxiliary fields.}
For visual arguments, the official scorer uses the role label together with the predicted bounding box to determine matching.
The optional \texttt{object\_name} field is retained for structural readability and to support model-side reasoning, but it is not required by the scorer and is not used as a matching key.

\noindent \textbf{Why this contract is shared.}
Using the same final output contract across all generative settings ensures that performance differences cannot be attributed to output-format freedom or privileged serialization choices, but instead reflect differences in reasoning and coordination under the same grounding and schema constraints.

\noindent \textbf{Normalization boundary.}
After role binding, triggers and textual arguments are aligned to minimal surface spans in $T$ for evaluation.
This extractive cleanup is evaluation-facing only: it does not alter event type, argument role, modality assignment, or event--argument membership.
Internal fields used by ECHO for intermediate reasoning---such as vertex identifiers, confidence scores, and operation rationales---are retained only in the MEHG/audit trail and are stripped from the final scored JSON.

\inputjsonbox{appendix/examples/output_event.json}{Event Extraction Example}

\subsection{Visual Context and Tool Augmentation}
\label{app:visual_tool_aug}

In the controlled comparison, the default vision tool,
Qwen3-VL-8B-Thinking, provides two forms of auxiliary visual evidence:
(i) a compact textual visual context $\mathcal{C}_I$ (e.g., scene summaries and object inventories), and
(ii) localized box proposals for candidate visual evidence.
When available, object names attached to these proposals are treated as auxiliary descriptors for readability and model-side reasoning.

Text-only LLMs receive the document text $T$ together with the tool-generated $\mathcal{C}_I$ and box proposals.
LVLMs receive the raw image $I$ together with the same auxiliary tool outputs.
Within ECHO, these tool outputs are used in two places:
Stage~I uses them to initialize the shared multimodal evidence inventory, and Stage~III uses event-conditioned visual queries together with localization consistency to bind visual roles.

This shared visual front-end reduces confounding from basic visual perception differences across generative methods.
At the same time, the choice of vision tool still matters:
as shown by the vision-tool ablation, a stronger tool improves argument-role performance in both visual and multimedia settings.
We therefore treat the default tool as a common grounding interface rather than a source of method-specific advantage.

\subsection{MEHG Representation and Identifiers}
\label{subsec:appendix-mehg}

ECHO maintains an explicit intermediate state as a Multimedia Event Hypergraph (MEHG).

\noindent \textbf{Vertices and IDs.}
The vertex set is $V = V_T \cup V_I$, where $V_T$ contains textual mention vertices and $V_I$ contains visual region vertices.
For unambiguous reference inside the MEHG, text vertices are indexed as \texttt{T1}, \texttt{T2}, \dots, visual region vertices as \texttt{O1}, \texttt{O2}, \dots, and event hyperedges as \texttt{HE1}, \texttt{HE2}, \dots.
When serialized MEHG states are shown for analysis, argument entries may additionally carry a \texttt{node\_id} field; every such \texttt{node\_id} must match a previously defined identifier in the corresponding node inventory.
These identifiers are internal bookkeeping fields used in the MEHG and audit trail, rather than part of the final scored JSON contract in Appendix~\ref{app:schema}.

\noindent \textbf{Hyperedge contents.}
Each hyperedge stores the trigger span $t_k$, event type $y_k^e$, a set of currently linked candidate vertices $V_k \subseteq V$, an optional role assignment $A_k$ (populated in Stage~III), and a confidence score $c_k$.
This representation makes intermediate event hypotheses inspectable across coordination rounds and supports audit-trail analysis of committed updates.

\noindent The example below shows an internal serialized MEHG state rather than the final evaluation JSON.

\inputjsonbox{appendix/examples/hypergraph.json}{Multimedia Event Hypergraph Example}

\subsection{Atomic Operation Schema and Audit Trail}
\label{subsec:appendix-ops}

Stage~II evolves the MEHG through a strictly defined set of atomic operations.
At each round, agents propose candidate updates to the current MEHG state, and only committed updates are recorded in the audit trail for traceability and post-hoc inspection.
Stage~II is restricted to topological edits: it updates event identity, candidate membership, and confidence, while deferring role assignment to Stage~III.

\noindent \textbf{Operation families.}
We define a fixed set of operations that includes proposing or revising an event hypothesis (\texttt{propose}/\texttt{revise} a hyperedge), removing a hypothesis (\texttt{drop}), linking or unlinking a vertex to a hyperedge (\texttt{link}/\texttt{unlink}), and adjusting confidence (\texttt{adjust\_confidence}).
Each operation includes a short natural-language rationale for transparency and qualitative analysis.

\noindent \textbf{Commit checks and conflict resolution.}
Before applying any operation, we perform lightweight structural checks to ensure that the operation is well-formed, references valid vertices or hyperedges, and satisfies basic schema constraints with respect to the current MEHG state.
When multiple surviving operations target the same structure within one round, we resolve conflicts deterministically before commit:
\texttt{drop} overrides other modifications to the same hyperedge;
\texttt{unlink} overrides \texttt{link} for the same node--hyperedge pair; and
conflicting \texttt{revise} operations defer to the Verifier.
Committed operations are recorded sequentially in the audit trail, which also acts as shared memory to discourage equivalent repeated updates across rounds.
Stage~II stops when no valid operation survives checking and conflict resolution.

\subsection{Prompt Templates (Abbreviated) for ECHO and Direct Prompting}
\label{app:prompts}

We use structured prompts that specify (i) the event and role inventories, (ii) the unified JSON contract (Appendix~\ref{app:schema}), and (iii) modality-integrity constraints.
To keep the appendix compact, we provide abbreviated templates with placeholders; the full prompts will be released with the codebase.

\noindent\textbf{Direct Prompting (LLM + Visual Tool Outputs).}
Text-only LLMs receive the document text $T$ along with the textual visual representation $\mathcal{C}_I$ (e.g., object inventory and image summary) produced by \texttt{Qwen3-VL-8B-Thinking}.

\inputtextbox{appendix/prompt/direct_llm.txt}{Prompt for Text-only LLMs}

\noindent\textbf{Direct Prompting (LVLM + Image + Visual Tool Outputs).}
LVLM baselines process the raw image $I$ alongside the text $T$. Crucially, to maintain a strictly controlled comparison, they are additionally provided with the exact same vision tool outputs ($\mathcal{C}_I$ and localized bounding box coordinates) as the text-only LLMs. They are constrained to the identical JSON output contract.

\inputtextbox{appendix/prompt/direct_lvlm.txt}{Prompt for LVLMs}

\noindent\textbf{ECHO.}
These prompts define each agent's role, available multimodal evidence, and the structured atomic operation format.

\inputtextbox{appendix/prompt/node_seeding.txt}{Node Seeding}

\inputtextbox{appendix/prompt/proposer.txt}{Proposer}

\inputtextbox{appendix/prompt/linker.txt}{Linker}

\inputtextbox{appendix/prompt/verifier.txt}{Verifier}

\noindent\textbf{Role Bind (textual vertices).}
The prompt below is used for textual role binding. Visual role candidates are obtained from the vision tool conditioned on the event hypothesis and then aligned to linked visual vertices by localization consistency, as described in Stage~III.

\inputtextbox{appendix/prompt/role_binder.txt}{Role Bind}

\inputtextbox{appendix/prompt/enhancer.txt}{Text Span Alignment}

\begin{table*}[t]
\centering
\caption{Detailed evaluation results of Direct Prompting LLMs/LVLMs. \textbf{DS} denotes DeepSeek models. The table reports Precision (P), Recall (R), and F1 scores across textual, visual, and multimedia settings. \best{Bold} and \second{underlined} values denote the best and second-best F1 scores in each column, respectively. All methods use the same default vision tool outputs generated by Qwen3-VL-8B-Thinking.}
\renewcommand{\arraystretch}{1.10}         

\begin{tabular}{@{} l 
    *{3}{r}*{3}{r} @{\hspace{0.8em}} % Textual 与 Visual 之间的间距
    *{3}{r}*{3}{r} @{\hspace{0.8em}} % Visual 与 Multimedia 之间的间距
    *{3}{r}*{3}{r} @{}}
\toprule
\multirow{3}{*}{\textbf{Method}} &
\multicolumn{6}{c}{\textbf{Textual Events}} &
\multicolumn{6}{c}{\textbf{Visual Events}} &
\multicolumn{6}{c}{\textbf{Multimedia Events}} \\
\cmidrule(lr){2-7} \cmidrule(lr){8-13} \cmidrule(lr){14-19}
& \multicolumn{3}{c}{Event Mention} & \multicolumn{3}{c}{Argument Role}
& \multicolumn{3}{c}{Event Mention} & \multicolumn{3}{c}{Argument Role}
& \multicolumn{3}{c}{Event Mention} & \multicolumn{3}{c}{Argument Role} \\
\cmidrule(lr){2-4} \cmidrule(lr){5-7}
\cmidrule(lr){8-10} \cmidrule(lr){11-13}
\cmidrule(lr){14-16} \cmidrule(lr){17-19}
& P & R & F1 & P & R & F1 & P & R & F1 & P & R & F1 & P & R & F1 & P & R & F1 \\
\midrule

% --- Data Rows ---
X-MTL
& 49.7 & 65.7 & 56.6 & 34.6 & 37.6 & \best{36.0} % Textual
& 73.1 & 70.3 & 71.7 & 33.2 & 31.3 & 32.2 % Visual
& 78.3 & 57.3 & 66.2 & 40.3 & 42.6 & \best{41.4} \\ % Multimedia

Qwen2.5-VL
& 48.0 & 38.8 & 42.9 & 10.3 & 13.9 & 11.8 
& 82.1 & 80.9 & \second{81.5} & 50.0 & 26.7 & 34.8 
& 34.8 & 35.0 & 34.9 & 28.0 & 18.1 & 22.0 \\

Qwen3-VL
& 56.9 & 49.4 & 52.9 & 10.3 & 17.1 & 12.9 
& 75.1 & 85.1 & 79.8 & 20.6 & 21.8 & 21.2 
& 61.7 & 61.7 & 61.7 & 21.1 & 24.6 & 22.7 \\

Qwen3-8B
& 62.5 & 43.8 & 51.5 & 14.2 & 16.9 & 15.5 
& 77.0 & 60.6 & 67.9 & 39.3 & 21.8 & 28.1 
& 29.0 & 32.0 & 30.5 & 19.4 & 18.6 & 19.0 \\

Qwen3-14B
& 67.1 & 44.8 & 53.7 & 14.7 & 16.6 & 15.6 
& 75.4 & 50.5 & 60.5 & 46.3 & 18.6 & 26.6 
& 60.0 & 65.0 & 62.4 & 25.8 & 21.6 & 23.5 \\

Qwen3-32B
& 62.5 & 56.3 & 59.2 & 13.8 & 21.0 & 16.7 
& 77.2 & 66.5 & 71.4 & 43.2 & 29.1 & 34.8 
& 51.1 & 67.0 & 58.0 & 23.2 & 24.0 & 23.6 \\

GPT-4o
& 66.0 & 45.5 & 53.9 & 12.8 & 14.3 & 13.5 
& 77.4 & 63.8 & 70.0 & 36.6 & 29.8 & 32.9 
& 65.1 & 77.3 & 70.7 & 14.8 & 26.4 & 18.9 \\

DS-V3(0324) % Abbreviated
& 58.3 & 63.3 & \second{60.7} & 11.5 & 20.1 & 14.6 
& 71.4 & 79.7 & 75.4 & 40.2 & 31.0 & 35.0 
& 56.3 & 79.2 & 65.9 & 13.9 & 28.4 & 18.7 \\

DS-V3.2 % Abbreviated
& 62.6 & 52.7 & 57.2 & 11.2 & 15.7 & 13.1 
& 74.3 & 81.4 & 77.7 & 43.3 & 33.9 & \second{38.0} 
& 55.6 & 79.9 & 65.6 & 23.9 & 30.2 & 26.7 \\

GPT-5
& 67.5 & 54.0 & 60.0 & 13.2 & 16.3 & 14.5 
& 81.5 & 80.2 & 80.9 & 39.0 & 39.9 & \best{39.5} 
& 71.1 & 81.8 & \best{76.1} & 25.5 & 33.8 & \second{29.1} \\

Gemini-3-Pro
& 57.8 & 83.0 & \best{68.1} & 14.5 & 32.4 & \second{20.1} 
& 75.7 & 91.4 & \best{82.8} & 23.0 & 26.2 & 24.5 
& 52.3 & 85.6 & 64.9 & 23.6 & 39.6 & 29.6 \\
\bottomrule
\end{tabular}
\label{tab:direct_prompting_full}
\end{table*}

\inputtextbox{appendix/prompt/synthesizer.txt}{Output}

% =========================================================
\section{Detailed Experimental Results and Further Discussion}
\label{app:extra_results}
This section reports the full precision/recall/F1 breakdown for the additional experiments discussed in the paper.
Unless otherwise noted, all results follow the official M2E2 evaluation protocol and are reported under the textual, visual, and multimedia settings for both Event Mention and Argument Role.

\subsection{Note on an Additional 2025 Baseline}
\label{sec:additional_2025_baselines}

Yu et al.~\cite{yu2025m2e2ke} propose an LLM knowledge-editing approach for M2E2. We do not include it in our numerical end-to-end comparison because its argument extraction stage uses oracle event types. By contrast, ECHO and the core baselines in this work must jointly discover event structures and role bindings without oracle event-type input, so a direct numerical comparison would not be methodologically aligned.

\subsection{Direct Prompting Baselines (P/R/F1)}
\label{app:prompt_full}

Table~\ref{tab:direct_prompting_full} provides the full results of Direct Prompting across all settings.

\noindent \textbf{Additional Analysis.}
Across both LLM and LVLM instantiations, scaling the backbone improves Event Mention more consistently than Argument Role.
Within the same model family, EM tends to benefit from larger capacity, whereas AR remains more constrained by faithful grounding and schema-consistent role binding.
This gap is especially pronounced in visual and multimedia settings: even under the same tool-augmented setup, single-pass Direct Prompting can yield competitive EM while remaining substantially weaker on AR, indicating that grounded role assignment is still brittle without explicit intermediate-state revision.

\subsection{Dialogue-Mediated Baseline Results (P/R/F1)}
\label{app:dialogue_baseline_details}

Table~\ref{tab:dialogue_baseline_results} presents the complete results of the dialogue-mediated baseline across all settings, using the same metrics and reporting conventions as above.

\noindent \textbf{Additional Analysis.}
Compared with Direct Prompting, dialogue-mediated coordination often improves grounded role extraction, indicating that iterative interaction can partially correct early mistakes.
However, these gains are not consistently monotonic with backbone strength.
This suggests that when intermediate state is carried implicitly in dialogue, stronger models can still exhibit unstable role binding or over-generation, which further motivates explicit structured state and operation-level validation used in ECHO.

\subsection{Ablation Study Details (P/R/F1)}
\label{app:ablation_details}

Table~\ref{tab:ablation_full} reports the full ablation results in the multimedia setting across different backbones.
Each variant removes one key component (Linker, Verifier, SpanAlign, or Link-then-Bind) to quantify its contribution under the same evaluation protocol.
For \emph{w/o Linker}, we adopt an all-to-all fallback that associates every candidate vertex with each event hypothesis, replacing negotiated relevance linking.

\begin{table*}[t]
\centering
\caption{Detailed evaluation results of the dialogue-mediated baseline. \textbf{DS} denotes DeepSeek models. The table reports Precision (P), Recall (R), and F1 scores across textual, visual, and multimedia settings. \best{Bold} and \second{underlined} values denote the best and second-best F1 scores in each column, respectively. This baseline uses the same default vision tool outputs generated by Qwen3-VL-8B-Thinking.}
\label{tab:dialogue_baseline_results}
{%

\begin{tabular}{@{} l
    *{3}{r}*{3}{r}@{\hspace{0.6em}}
    *{3}{r}*{3}{r}@{\hspace{0.6em}}
    *{3}{r}*{3}{r} @{}}
\toprule
\multirow{3}{*}{\textbf{Method}} &
\multicolumn{6}{c}{\textbf{Textual Events}} &
\multicolumn{6}{c}{\textbf{Visual Events}} &
\multicolumn{6}{c}{\textbf{Multimedia Events}} \\
\cmidrule(lr){2-7} \cmidrule(lr){8-13} \cmidrule(lr){14-19}
& \multicolumn{3}{c}{Event Mention} & \multicolumn{3}{c}{Argument Role}
& \multicolumn{3}{c}{Event Mention} & \multicolumn{3}{c}{Argument Role}
& \multicolumn{3}{c}{Event Mention} & \multicolumn{3}{c}{Argument Role} \\
\cmidrule(lr){2-4} \cmidrule(lr){5-7}
\cmidrule(lr){8-10} \cmidrule(lr){11-13}
\cmidrule(lr){14-16} \cmidrule(lr){17-19}
& P & R & F1 & P & R & F1 & P & R & F1 & P & R & F1 & P & R & F1 & P & R & F1 \\
\midrule

X-MTL
& 49.7 & 65.7 & 56.6 & 34.6 & 37.6 & \best{36.0}
& 73.1 & 70.3 & 71.7 & 33.2 & 31.3 & 32.2
& 78.3 & 57.3 & \second{66.2} & 40.3 & 42.6 & \second{41.4} \\

Qwen3-8B
& 50.3 & 65.6 & 56.9 & 20.5 & 41.2 & 27.4
& 65.0 & 84.5 & 73.5 & 28.2 & 57.6 & 37.9
& 44.7 & 71.4 & 55.0 & 26.3 & 61.0 & 36.8 \\

Qwen3-32B
& 51.3 & 67.7 & \second{58.3} & 21.2 & 43.5 & \second{28.6}
& 67.1 & 83.3 & \second{74.3} & 46.7 & 52.9 & \second{49.6}
& 48.4 & 77.6 & 59.6 & 43.4 & 42.7 & \best{43.0} \\

DS-V3.2
& 59.0 & 69.0 & \best{63.6} & 21.7 & 39.4 & 28.0
& 74.0 & 84.6 & \best{78.9} & 52.6 & 50.9 & \best{51.7}
& 60.0 & 83.8 & \best{69.9} & 30.0 & 55.2 & 38.8 \\

\bottomrule
\end{tabular}
}
\end{table*}

\begin{table*}[t]
\centering
\caption{Full ablation results in the multimedia setting (P/R/F1). We evaluate ECHO across different backbones: Qwen3 (8B, 32B) and DeepSeek-V3.2. The ablations test the impact of removing the Linker (relevance linking), Verifier, SpanAlign (Stage~III text span alignment), and Link-then-Bind. All variants use the same default vision tool outputs.}
\resizebox{\textwidth}{!}{
\begin{tabular}{
    l
    *{6}{S[table-format=2.1]} % Qwen3-8B
    *{6}{S[table-format=2.1]} % Qwen3-32B
    *{6}{S[table-format=2.1]} % DS-v3.2
}
\toprule
\multirow{2}{*}{\textbf{Method}} &
\multicolumn{6}{c}{\textbf{Qwen3-8B}} &
\multicolumn{6}{c}{\textbf{Qwen3-32B}} &
\multicolumn{6}{c}{\textbf{DeepSeek-v3.2}} \\
\cmidrule(lr){2-7} \cmidrule(lr){8-13} \cmidrule(lr){14-19}
& \multicolumn{3}{c}{Event Mention} & \multicolumn{3}{c}{Argument Role}
& \multicolumn{3}{c}{Event Mention} & \multicolumn{3}{c}{Argument Role}
& \multicolumn{3}{c}{Event Mention} & \multicolumn{3}{c}{Argument Role} \\
\cmidrule(lr){2-4} \cmidrule(lr){5-7} 
\cmidrule(lr){8-10} \cmidrule(lr){11-13} 
\cmidrule(lr){14-16} \cmidrule(lr){17-19}
& {P} & {R} & {F1} & {P} & {R} & {F1} 
& {P} & {R} & {F1} & {P} & {R} & {F1}
& {P} & {R} & {F1} & {P} & {R} & {F1} \\
\midrule
\textbf{Full Model}      
& 70.6 & 69.9 & 70.2 & 47.0 & 58.9 & 52.3   % Qwen-8B
& 72.5 & 72.5 & 72.5 & 49.8 & 61.2 & 54.9   % Qwen-32B
& 75.1 & 75.1 & 75.1 & 47.2 & 65.8 & 55.0 \\ % DS-v3.2
\addlinespace[0.35em]
- w/o Linker     
& 69.9 & 69.9 & 69.9 & 49.7 & 45.8 & 47.7 
& 70.6 & 70.6 & 70.6 & 42.1 & 37.3 & 39.6 
& 74.8 & 74.8 & 74.8 & 45.2 & 40.3 & 42.6 \\
- w/o Verifier   
& 68.9 & 68.9 & 68.9 & 46.0 & 56.8 & 50.8 
& 71.8 & 71.8 & 71.8 & 44.8 & 50.8 & 47.6 
& 74.8 & 74.8 & 74.8 & 46.6 & 56.4 & 51.0 \\
- w/o SpanAlign  
& 64.7 & 64.7 & 64.7 & 44.5 & 54.9 & 49.1 
& 67.3 & 67.3 & 67.3 & 38.5 & 48.5 & 42.9 
& 74.8 & 74.8 & 74.8 & 38.6 & 47.7 & 42.6 \\
- w/o Link-then-Bind        
& 67.4 & 67.0 & 67.2 & 31.3 & 17.5 & 22.4 
& 69.9  & 69.9 & 69.9 & 34.6 & 18.1 & 23.7 
& 76.4 & 76.4 & 76.4 & 28.2 & 20.9 & 24.0 \\
\bottomrule
\end{tabular}
}
\label{tab:ablation_full}
\end{table*}

\noindent \textbf{Additional Analysis.}
The ablations show complementary contributions beyond aggregate F1 changes.
Removing Link-then-Bind consistently lowers AR, indicating that early role conditioning makes Stage~II overly conservative before the relevance topology stabilizes; although a slight EM gain can appear for a stronger backbone, it is outweighed by the larger loss in argument-role accuracy.
Removing the Linker mainly reduces AR coverage, suggesting that negotiated relevance linking is important for surfacing a sufficiently rich candidate set for downstream binding.
The Verifier becomes more important as the backbone strengthens, consistent with stronger models generating a broader and potentially noisier hypothesis space that benefits from explicit validation and pruning.
SpanAlign primarily addresses extractive constraints by correcting span-level mismatches, and its impact is larger when the backbone is less reliable at producing evaluation-compatible text fields.

% =========================================================

\section{Qualitative Analysis and Error Analysis}
\label{app:qualitative}

\subsection{Case Study with Audit Trails}
\label{app:case_study}

We consider a representative M2E2 document in which the text states that \emph{militants} \emph{ride in vehicles toward Iraq}, while the paired image provides a visually compatible vehicle region.
ECHO first performs Stage~I node seeding to build the initial MEHG inventory, and then carries out Stage~II negotiation under the Link-then-Bind schedule: candidate mentions and regions are linked to event hypotheses before any fine-grained semantic role is finalized.
In this example, the Proposer considers a weaker \texttt{Conflict:Attack} hypothesis and a stronger \texttt{Movement:Transport} hypothesis.
The Verifier favors the transport hypothesis because the trigger, destination mention, and vehicle evidence are mutually supportive across modalities, as summarized in Table~\ref{tab:audit_excerpt}.
Stage~III then binds roles and consolidates the transport event as the final prediction, while the weaker attack alternative is suppressed.

\begin{table}[t]
\centering
\caption{Audit-trail excerpt for the representative transport example discussed in this subsection. Representative committed updates are shown; \textsc{P}/\textsc{L}/\textsc{V}/\textsc{B} denote Proposer, Linker, Verifier, and Role Binder, respectively, and \textit{adj.\ conf.} denotes \texttt{adjust\_confidence}. Additional linking steps are omitted for brevity.}
\label{tab:audit_excerpt}
\begin{tabular}{@{}p{0.55cm}p{0.55cm}p{1.55cm}p{4.55cm}@{}}
\toprule
Step & Ag. & Op. & Target / Outcome \\
\midrule
1 & P & propose & $\epsilon_{\textsc{atk}}$: \texttt{Conflict:Attack} \\
2 & P & propose & $\epsilon_{\textsc{trans}}$: \texttt{Movement:Transport} \\
3 & L & link & $\epsilon_{\textsc{trans}} \leftarrow$ \\
  &   &      & \hspace{0.6em}\{ \textit{militants}, \textit{vehicles}, \textit{Iraq}, \textit{vehicle region}~(\textsc{img}) \} \\
4 & V & adj.\ conf. & $\epsilon_{\textsc{atk}}: \varnothing \rightarrow 0.70 \rightarrow 0.60$ (insufficient multimodal support) \\
5 & V & adj.\ conf. & $\epsilon_{\textsc{trans}}: \varnothing \rightarrow 0.90 \rightarrow 0.95$ (strong cross-modal support) \\
6 & B & bind & Bind roles for $\epsilon_{\textsc{trans}}$; discard unsupported \textit{weapons} \\
\bottomrule
\end{tabular}
\end{table}

\noindent \textbf{Consolidated output.}
The retained event is \texttt{Movement:Transport} with trigger \textit{ride}.
Its final bindings include \textit{militants} (\texttt{Agent}), \textit{vehicles} (\texttt{Vehicle}), and \textit{Iraq} (\texttt{Destination}), together with one aligned visual vehicle region.
The candidate \textit{weapons} is not retained because it receives no sufficiently supported role during Stage~III binding.

\begin{table}[t]
\centering
\caption{AR error counts for Qwen3-32B under ECHO on M2E2.}
\label{tab:ar_error_qwen32b_appendix}
\begin{tabular}{lccc}
\toprule
\textbf{AR error type} & \textbf{Textual} & \textbf{Visual} & \textbf{Multimedia} \\
\midrule
Spurious                         & 540 & 0   & 180 \\
Span mismatch                    & 552 & 0   & 181 \\
Role misassignment               & 60  & 12  & 54  \\
No-gold event type               & 768 & 111 & 208 \\
Localization error               & 0   & 173 & 400 \\
\midrule
\textbf{Total}                   & 1920 & 296 & 1023 \\
\bottomrule
\end{tabular}
\end{table}

\begin{table}[h]
\centering
\caption{Normalized string relations between predicted and gold text arguments for Qwen3-32B under ECHO.}
\label{tab:q32_spanrel}
\begin{tabular}{lrr}
\toprule
\textbf{String relation} & \textbf{Textual} & \textbf{Multimedia} \\
\midrule
No-gold      & 1536 & 184 \\
Exact        & 987  & 282 \\
None         & 994  & 329 \\
Contained-by & 65   & 20  \\
Contains     & 36   & 14  \\
Overlap      & 3    & 0   \\
\bottomrule
\end{tabular}
\end{table}

\begin{table}[h]
\centering
\caption{Most frequent roles involved in Argument Role misassignments for Qwen3-32B under ECHO.}
\label{tab:q32_roleconf}
\setlength{\tabcolsep}{4.2pt}
\renewcommand{\arraystretch}{1.08}
\begin{tabular}{ll>{\raggedright\arraybackslash}p{0.5\linewidth}}
\toprule
\textbf{Setting} & \textbf{Source} &
\makecell[l]{\textbf{Most frequent}\\\textbf{misassigned roles}} \\
\midrule
Textual     & Text &
Agent (18), Target (10), Attacker  (9), Artifact (7), Place (6) \\
Visual      & Box  &
Entity (5), Agent (5), Recipient(1), Artifact (1) \\
Multimedia  & Text &
Target (7), Place (6), Attacker(3), Agent (2), Entity (1) \\
Multimedia  & Box  &
Instrument (6), Artifact (5), Target (5), Agent (4), Entity (4) \\
\bottomrule
\end{tabular}
\end{table}

\subsection{Error Breakdown under ECHO}
\label{app:error_breakdown_qwen32b}

We report a finer-grained error breakdown for Qwen3-32B under ECHO on M2E2, covering the textual, visual, and multimedia settings.
Text arguments are matched to gold after the same evaluation-facing normalization described in Appendix~\ref{app:schema}, while image arguments follow the official box-matching protocol.

\noindent \textbf{AR error taxonomy.}
We categorize Argument Role errors into five types:
1) \textbf{Spurious}: predicted arguments not supported by any gold role in the instance;
2) \textbf{Span mismatch}: plausible role but generated text does not match gold after normalization;
3) \textbf{Role misassignment}: correct content (or matched box) but incorrect role label;
4) \textbf{No-gold event type}: arguments attached to predicted event types without gold arguments in the instance; and
5) \textbf{Localization error}: image arguments that fail to match any gold box under the official scorer criterion.

\begin{table}[t]
\centering
\caption{EM error counts for Qwen3-32B under ECHO.}
\label{tab:q32_emerr}
\setlength{\tabcolsep}{4.8pt}
\renewcommand{\arraystretch}{1.08}
\begin{tabular}{lccc}
\toprule
\textbf{Setting} & \textbf{Spurious} & \textbf{Missing} & \textbf{Trigger mismatch} \\
\midrule
Textual    & 469 & 371 & 129 \\
Visual     & 37  & 37  & 0   \\
Multimedia & 85  & 85  & 52  \\
\bottomrule
\end{tabular}
\end{table}

\begin{table}[h]
\centering
\caption{Argument Role (AR) output statistics under ECHO on M2E2. \textbf{DS} denotes DeepSeek models. \textbf{Matched} counts are computed using the same official AR matching protocol as in the main evaluation. \textbf{OverGen} denotes the over-generation ratio (\textit{Pred}/\textit{Gold}).}
\label{tab:ar_output_stats_echo}
\resizebox{\columnwidth}{!}{%
\begin{tabular}{lrrrrrr}
\toprule
\multicolumn{7}{c}{\textbf{Textual} (Gold = 1659)} \\
\cmidrule(lr){1-7}
Model & Pred & Matched & FP & P(\%) & R(\%) & OverGen \\
\midrule
Qwen3-8B        & 2771 & 893  & 1878 & 32.2 & 53.8 & 1.67 \\
Qwen3-14B       & 2620 & 881  & 1739 & 33.6 & 53.1 & 1.58 \\
Qwen3-32B       & 2853 & 933  & 1920 & 32.7 & 56.2 & 1.72 \\
DS-V3.2   & 2936 & 949  & 1987 & 32.3 & 57.2 & 1.77 \\
GPT-5           & 3796 & 1033 & 2763 & 27.2 & 62.3 & 2.29 \\
\midrule
\multicolumn{7}{c}{\textbf{Visual} (Gold = 688)} \\
\cmidrule(lr){1-7}
Model & Pred & Matched & FP & P(\%) & R(\%) & OverGen \\
\midrule
Qwen3-8B        & 670 & 413 & 257 & 61.6 & 60.0 & 0.97 \\
Qwen3-14B       & 662 & 410 & 252 & 61.9 & 59.6 & 0.96 \\
Qwen3-32B       & 641 & 396 & 245 & 61.8 & 57.6 & 0.93 \\
DS-V3.2   & 638 & 400 & 238 & 62.7 & 58.1 & 0.93 \\
GPT-5           & 649 & 401 & 248 & 61.8 & 58.3 & 0.94 \\
\midrule
\multicolumn{7}{c}{\textbf{Multimedia} (Gold = 1460)} \\
\cmidrule(lr){1-7}
Model & Pred & Matched & FP & P(\%) & R(\%) & OverGen \\
\midrule
Qwen3-8B        & 1831 & 860 & 971  & 47.0 & 58.9 & 1.25 \\
Qwen3-14B       & 1770 & 861 & 909  & 48.6 & 59.0 & 1.21 \\
Qwen3-32B       & 1792 & 893 & 899  & 49.8 & 61.2 & 1.23 \\
DS-V3.2   & 2033 & 960 & 1073 & 47.2 & 65.8 & 1.39 \\
GPT-5           & 2069 & 927 & 1142 & 44.8 & 63.5 & 1.42 \\
\bottomrule
\end{tabular}%
}
\end{table}

\begin{table*}[t]
\centering
\caption{Interaction and computation cost under an idealized schedule.
We assume $T_{\max}{=}2$ rounds with no early stopping and a single extracted event, and report (left) the number of main-model and vision-tool calls and (right) token usage (input/output/total) for \textsc{ECHO} and the dialogue-mediated baseline across textual, visual, and multimedia settings.
This table is intended as an illustrative cost accounting under the controlled protocol rather than an average empirical runtime summary.
In practice, call counts can be lower due to early stopping ($T_{\mathrm{used}}<T_{\max}$) and vary with the number of extracted events.}
\label{tab:cost_profile}
\begin{tabular}{l ccc ccc | ccc ccc}
\toprule
% --- 第一层表头：两大核心指标 ---
\multirow{3}{*}{\textbf{Setting}} & 
\multicolumn{6}{c|}{\textbf{Interaction Cost (Number of Calls)}} & 
\multicolumn{6}{c}{\textbf{Computation Cost (Token Usage)}} \\
\cmidrule(lr){2-7} \cmidrule(lr){8-13}

% --- 第二层表头：模型对比 ---
& \multicolumn{3}{c}{\textbf{Dialogue-mediated}} & \multicolumn{3}{c|}{\textbf{ECHO (Ours)}} & 
\multicolumn{3}{c}{\textbf{Dialogue-mediated}} & \multicolumn{3}{c}{\textbf{ECHO (Ours)}} \\
\cmidrule(lr){2-4} \cmidrule(lr){5-7} \cmidrule(lr){8-10} \cmidrule(lr){11-13} 

% --- 第三层表头：细分项 ---
& Main & Vis & \textbf{Total} & Main & Vis & \textbf{Total} 
& Input & Output & \textbf{Total} & Input & Output & \textbf{Total} \\
\midrule

% --- Text-only ---
Textual
& 8 & 0 & 8 
& 9 & 0 & 9 
& 29{,}110 & 6{,}065 & 35{,}175 
& 14{,}416 & 13{,}141 & \textbf{27{,}557} \\

% --- Visual ---
Visual
& 7 & 2 & 9 
& 8 & 2 & 10 
& 42{,}384 & 7{,}996 & 50{,}380 
& 28{,}191 & 16{,}812 & \textbf{45{,}003} \\

% --- Multimedia ---
Multimedia
& 8 & 2 & 10 
& 9 & 2 & 11 
& 54{,}460 & 13{,}123 & 67{,}582 
& 34{,}695 & 22{,}210 & \textbf{56{,}905} \\

\bottomrule
\end{tabular}
\end{table*}

\noindent \textbf{Overall AR distribution.}
Table~\ref{tab:ar_error_qwen32b_appendix} reports the error counts by type.
Consistent with the overall analysis above, textual and multimedia AR errors are dominated by over-generation and span mismatch, whereas visual-involved settings remain strongly constrained by localization quality.
The following tables further break this picture down into text-span discrepancy patterns, frequent role confusions, and event-mention error types.

\noindent \textbf{Text span discrepancy profile.}
To better characterize span-related behavior, we compute the normalized string relation between each predicted text argument and its best-matched gold text, if any.
Table~\ref{tab:q32_spanrel} reports the resulting distribution over all predicted text arguments after normalization; therefore, \textit{Exact} includes already-correct matches and the counts do not coincide with the AR error totals in Table~\ref{tab:ar_error_qwen32b_appendix}.
Among the non-exact cases, \textit{None} and \textit{No-gold} dominate, while containment/overlap cases are relatively rare.
This suggests that many text-side failures arise from non-extractive or semantically off-target generations rather than minor boundary shifts.

\noindent \textbf{Frequent roles involved in misassignments.}
Role misassignments concentrate on a small set of frequent schema roles.
Table~\ref{tab:q32_roleconf} lists the most common roles appearing in role-misassignment cases by setting and source, showing that text-side errors are concentrated in frequent participant/location roles, whereas box-side errors are concentrated in visually grounded roles.

\noindent \textbf{EM error analysis.}
For Event Mention, we categorize errors into \textbf{Spurious} (predicted event not matched to any gold event), \textbf{Missing} (gold event not predicted), and \textbf{Trigger mismatch} (event type matches, but the extracted trigger string differs after normalization).
Table~\ref{tab:q32_emerr} reports the corresponding counts.
As expected, trigger mismatch appears only in text-involved settings, whereas the visual setting contains no such errors because visual event mention scoring does not rely on textual triggers.

\noindent \textbf{Summary.}
Overall, the error profile is consistent with the preceding analysis:
Argument Role errors are dominated by over-generation and imperfect grounding, whereas Event Mention errors mainly come from spurious and missing events.
The next subsection complements this view with model-level output statistics.

\subsection{Output Statistics}
\label{app:output_stats}

Table~\ref{tab:ar_output_stats_echo} reports model-level Argument Role output statistics under ECHO.
Consistent with the main results, the textual setting shows clear over-generation of role bindings (OverGen $> 1$), which contributes to many false positives and lower precision.
By contrast, the visual setting is much closer to balanced generation (OverGen near $1$), while the multimedia setting remains moderately over-generated.

% =========================================================
\section{Efficiency Notes}
\label{app:efficiency}

\subsection{Call and Token Profile}
\label{app:token_profile}

Table~\ref{tab:cost_profile} summarizes interaction cost (calls) and computation cost (token usage) under an idealized schedule. For reference, Direct Prompting is the cheapest controlled baseline: in the visual and multimedia settings it uses one vision-tool call plus one main-model extraction call (two calls in total), and in the textual setting it reduces to a single main-model call.
Under the same round-budget assumption, \textsc{ECHO} incurs one additional main-model call relative to the dialogue-mediated baseline because it includes a dedicated Stage~III role-binding step, whereas the dialogue-mediated baseline uses a single final consolidator.
\textsc{ECHO} is therefore more expensive than Direct Prompting by design, because it adds iterative negotiation and a separate final binding stage.
Despite this extra cost, \textsc{ECHO} uses fewer total tokens than the dialogue-mediated baseline across settings, consistent with coordinating through compact state operations rather than accumulating long dialogue histories.

\end{document}

%% file: body/0-abs.tex
Multimedia event extraction (M2E2) aims to predict triggers, ground arguments across text and images, and assemble them into schema-consistent event records. 
Recent LLM-based approaches have shown strong potential for M2E2, but their intermediate event hypotheses often remain implicit, and event-argument linking is still tightly coupled with role binding. This leaves little opportunity to inspect or revise intermediate event hypotheses and makes predictions brittle to early errors.
To bridge this gap, we present \textbf{ECHO}, a multi-agent framework that reframes M2E2 as iterative refinement over an explicit \textbf{Multimedia Event Hypergraph} (MEHG).
Instead of relying on implicit linear generation, ECHO performs auditable atomic updates over a shared hypergraph, making intermediate event structures explicit and revisable.
Furthermore, we introduce a \textbf{Link-then-Bind} strategy that decouples event--argument linking from role binding, reducing premature semantic commitment during structured prediction. 
Extensive experiments on the M2E2 benchmark show that ECHO outperforms state-of-the-art approaches, achieving gains of 7.3 and 15.5 F1 points on event mention and argument role, respectively.